\documentclass{article}

\usepackage{amsmath}
\usepackage{amsfonts}
\usepackage{microtype}
\usepackage{graphicx}
\usepackage[center]{subfigure}
\usepackage{booktabs} % for professional tables

\usepackage[breaklinks=true]{hyperref}

\usepackage[accepted]{icml2022}

\icmltitlerunning{Variational Sparse Coding with Learned Thresholding}

\begin{document}

\twocolumn[
\icmltitle{Variational Sparse Coding with Learned Thresholding}

% It is OKAY to include author information, even for blind
% submissions: the style file will automatically remove it for you
% unless you've provided the [accepted] option to the icml2021
% package.

% List of affiliations: The first argument should be a (short)
% identifier you will use later to specify author affiliations
% Academic affiliations should list Department, University, City, Region, Country
% Industry affiliations should list Company, City, Region, Country

% You can specify symbols, otherwise they are numbered in order.
% Ideally, you should not use this facility. Affiliations will be numbered
% in order of appearance and this is the preferred way.
\icmlsetsymbol{equal}{*}

\begin{icmlauthorlist}
\icmlauthor{Kion Fallah}{to}
\icmlauthor{Christopher J. Rozell}{to}
\end{icmlauthorlist}

\icmlaffiliation{to}{ML@GT, Georgia Institute of Technology, Atlanta, Georgia}

\icmlcorrespondingauthor{Kion Fallah}{kion@gatech.edu}

% You may provide any keywords that you
% find helpful for describing your paper; these are used to populate
% the "keywords" metadata in the PDF but will not be shown in the document
\icmlkeywords{Machine Learning, ICML, Sparse Coding, Variational Inference}

\vskip 0.3in
]

% this must go after the closing bracket ] following \twocolumn[ ...

% This command actually creates the footnote in the first column
% listing the affiliations and the copyright notice.
% The command takes one argument, which is text to display at the start of the footnote.
% The \icmlEqualContribution command is standard text for equal contribution.
% Remove it (just {}) if you do not need this facility.

%\printAffiliationsAndNotice{}  % leave blank if no need to mention equal contribution
\printAffiliationsAndNotice{} % otherwise use the standard text.

\begin{abstract}
Sparse coding strategies have been lauded for their parsimonious representations of data that leverage low dimensional structure. However, inference of these codes typically relies on an optimization procedure with poor computational scaling in high-dimensional problems. For example, sparse inference in the representations learned in the high-dimensional intermediary layers of deep neural networks (DNNs) requires an iterative minimization to be performed at each training step. As such, recent, quick methods in variational inference have been proposed to infer sparse codes by learning a distribution over the codes with a DNN. In this work, we propose a new approach to variational sparse coding that allows us to learn sparse distributions by thresholding samples, avoiding the use of problematic relaxations. We first evaluate and analyze our method by training a linear generator, showing that it has superior performance, statistical efficiency, and gradient estimation compared to other sparse distributions. We then compare to a standard variational autoencoder using a DNN generator on the Fashion MNIST and CelebA datasets.
\end{abstract}

\begin{figure}
    \includegraphics[scale=0.54]{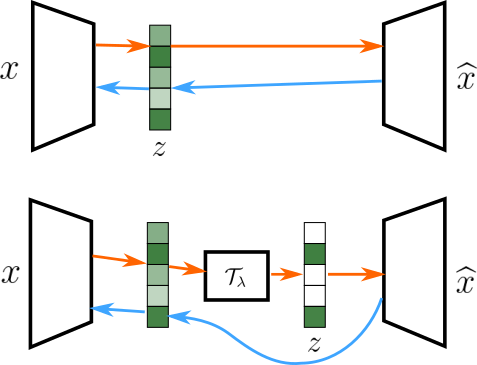}
    \centering
    \caption{Comparison of standard black-box variational inference and our proposed approach for variational sparse coding. Orange arrows depict a forward pass, blue arrows depict automatic differentiation. \textit{Top:} Sampling from standard variational inference approaches result in all features non-zero. \textit{Bottom:} Our approach incorporates sparsity via a shifted soft-threshold function. We utilize a straight-through estimator, skipping the gradient of the shifted soft-threshold, for exact sparsity without numerical instability during training.}
    %\vspace{-1.0em}
    \label{fig:intro_fig}
\end{figure}

\section{Introduction}
\label{intro}
Variational inference has become a ubiquitous tool in unsupervised learning of a distribution of latent features. These variational distributions can offer approximations in cases where inference over a true distribution is computationally expensive. Once inference is performed, latent features can be used for a variety of machine learning tasks, such as summarizing a dataset or training a generative model.
%The efficacy of these methods have been demonstrated in tasks including generative modeling, causal inference, and computational neuroscience.
The structure and statistical properties of latent features depend on the practitioner's choice of a prior distribution. Sparse distributions, in which only a few features are non-zero for each input data sample, have been favored for encouraging statistically efficient representations, especially when input data has low-dimensional structure \cite{olshausen_sparse_2004, elad_sparse_2010}.

To perform inference in sparse models with low computational cost, recent black-box variational inference (BBVI) \cite{ranganath_black_nodate} methods have been proposed to learn distribution parameters with DNNs using automatic differentiation \cite{kingma_auto-encoding_2014, rezende_stochastic_2014}. Unfortunately, these approaches either do not explicitly learn sparse features \cite{barello_sparse-coding_2018} or rely on relaxations that can lead to poor gradient estimation during training \cite{tonolini_variational_2020}.

Motivated by the success of soft-thresholding in iterative optimization procedures \cite{daubechies_iterative_2003, donoho_ideal_1994}, we propose a BBVI method to learn sparse distributions by thresholding samples drawn from a Laplacian or Gaussian distribution. We analytically show that thresholded samples from the former are identically distributed to a Spike-and-Slab distribution, giving practitioners control over the extent of sparsity by adjusting a threshold hyper-parameter. In cases where the degree of sparsity is not known beforehand, we also propose a technique to learn a distribution on the threshold parameter. To train our inference network, we apply a straight-through estimator \cite{bengio_estimating_2013, oord_neural_2018}, leading to favorable training stability and gradient estimation. Finally, we propose a new sampling procedure that encourages feature reuse, encouraging the generator to learn more diverse features. We showcase the performance of our method compared to other inference strategies by training and analyzing a linear generator on whitened image patches \cite{olshausen_emergence_1996} and a DNN generator on the Fashion MNIST \cite{https://doi.org/10.48550/arxiv.1708.07747} and CelebA \cite{liu2015faceattributes} datasets \footnote{Code available at: \href{https://github.com/kfallah/variational-sparse-coding}{\texttt{https://github.com/kfallah/\\variational-sparse-coding}}.}.

\section{Related Work}

\subsection{MAP Estimate/Regressive Inference}
Sparsity models have a long history in statistical modeling with methods such as LASSO regularization \cite{tibshirani1996regression}. These models often infer sparse latent features via a maximum a posteriori (MAP) estimate, requiring an iterative optimization procedure to be solved. The seminal work of \cite{olshausen_emergence_1996} proposes sparse codes as a means of unsupervised learning of a linear generator. Once trained, columns of the generator qualitatively resemble the receptive fields of mammalian cortical cells. Although methods exist to solve this optimization in discrete \cite{beck_fast_2009, yang_fast_2012} and continuous \cite{rozell_sparse_2008} time, their computational cost often scales poorly with dimensionality, making them prohibitive to use in modern deep learning settings.

An alternative approach for inference has been to use DNNs to regress sparse codes for given input data. One method ``unrolls'' iterations of the ISTA algorithm \cite{gregor_learning_nodate}. These methods are limited for unsupervised learning since they require ground truth codes as supervision during training.

\subsection{Variational Inference}
Early variational inference approaches applied exponential families for sparse coding, using iterative procedures to fit a variational posterior distribution \cite{girolami_variational_2001,seeger_bayesian_2008}. Later works explored variational methods for Spike-and-Slab models \cite{goodfellow_large-scale_nodate,sheikh_select-and-sample_2016}, using approximations to analytic solutions for faster inference. None of these approaches effectively scale inference to high-dimensional DNN representations.

The proposal of BBVI, where DNNs are used to estimate parameters of a variational posterior distribution, has led to a leap in the computational efficiency of variational inference \cite{kingma_auto-encoding_2014, rezende_stochastic_2014}. As such, many recent methods have been proposed to train DNN inference networks with various prior distributions. These include Laplacian \cite{barello_sparse-coding_2018}, Spike-and-Slab \cite{tonolini_variational_2020}, and Beta-Bernoulli \cite{singh_structured_nodate} distributions. Other work has incorporated sparsity through hierarchical posterior distributions \cite{salimans_structured_2016}, evolutionary variational algorithms \cite{JMLR:v23:20-233}, or group-sparsity in connections to generator networks \cite{ainsworth_oi-vae_nodate, moran_identifiable_2021}. We refer the reader to \cite{zhang_advances_2018} for a review of variational inference.

\subsection{Estimating the Variational Bound}
Various sampling approaches have been proposed to estimate the variational bound. \cite{cremer_reinterpreting_2017} proposes a sampling procedure based on a tighter bound on the data likelihood introduced in \cite{burda_importance_2016}. Counter-intuitively, using this tighter bound in training leads to a reduced signal-to-noise ratio in the inference network gradient \cite{rainforth_tighter_2019}. In \cite{grover_variational_2014}, the authors apply rejection sampling using a computed acceptance probability for each sample. A later work \cite{bauer_resampled_nodate} learns this acceptance probability with an additional DNN. Other work uses an alternative gradient estimator in BBVI which ignores certain terms during automatic differentiation to reduce variance \cite{roeder_sticking_2017}, with a bias-free estimator proposed by \cite{tucker_doubly_2018}.

\section{Methods}
\subsection{Black-box Variational Inference}
The capability of DNNs as universal function approximators has been recently applied to learn complex distributions. Given a dataset of training samples $[\mathbf{x}^1, \dots, \mathbf{x}^N]^T \in \mathbb{R}^{N \times D}$ from target density $p(\mathbf{x})$, BBVI can be applied to train an inference network $q_{\boldsymbol{\phi}}(\mathbf{z} \mid \mathbf{x})$ to learn a distribution over latent features $\mathbf{z}^{k} \in \mathbb{R}^{d}$. This distribution can be applied in various machine learning tasks, such as training a generator $p_{\boldsymbol{\theta}}(\mathbf{x} \mid \mathbf{z})$. To do this, one may employ the variational lower bound (ELBO) to maximize the marginal likelihood over the training samples \cite{jordan_introduction_1998}:
\begin{align}
  \log p_{\boldsymbol{\theta}}(\mathbf{x}) & = \log \mathbb{E}_{p(\mathbf{z})} \left[ p_{\boldsymbol{\theta}}(\mathbf{x} \mid \mathbf{z})\right] \nonumber \\
  & =  \log \mathbb{E}_{p(\mathbf{z})} \left[ \frac{q_{\boldsymbol{\phi}}(\mathbf{z} \mid \mathbf{x})}{q_{\boldsymbol{\phi}}(\mathbf{z} \mid \mathbf{x})} p_{\boldsymbol{\theta}}(\mathbf{x} \mid \mathbf{z})\right] \nonumber \\
  & \geq  \mathbb{E}_{q_{\boldsymbol{\phi}}} \left[\log p_{\boldsymbol{\theta}}(\mathbf{x} \mid \mathbf{z})\right] -  D_{KL}\bigl(q_{\boldsymbol{\phi}}(\mathbf{z}\mid\mathbf{x}) ~||~ p(\mathbf{z})\bigr) \nonumber \\
  & = \mathcal{L}\left(\boldsymbol{\theta}, \boldsymbol{\phi} ; \mathbf{x}\right).
  \label{eq:elbo}
\end{align}

Training under this bound requires the practitioner to select a prior $p(\mathbf{z})$ and a distribution family for the posterior $q_{\boldsymbol{\phi}}(\mathbf{z} \mid \mathbf{x})$. In this work, we follow the common assumption of independence for both, $p(\mathbf{z}) = \prod_{i=1}^d p(z_i), q_{\boldsymbol{\phi}}(\mathbf{z} \mid \mathbf{x}) = \prod_{i=1}^d q_{\boldsymbol{\phi}}(z_i \mid \mathbf{x})$. For each prior distribution we consider, we will use the same distribution family for our variational posterior.

Parameters $(\boldsymbol{\phi}, \boldsymbol{\theta})$ can be trained via a stochastic version of the variational expectation maximization algorithm \cite{neal_view_1998, barello_sparse-coding_2018}. Using batched training samples $\mathbf{x}^k$, this algorithm iterates in two steps from initializations $(\boldsymbol{\phi}^{(0)}, \boldsymbol{\theta}^{(0)})$:
\begin{enumerate}
  \item{Expectation: $\boldsymbol{\phi}^{(t)} = \arg \max_{\boldsymbol{\phi}} \mathcal{L}\left(\boldsymbol{\theta}^{(t-1)}, \boldsymbol{\phi} ; \mathbf{x}^k\right)$}
  \item{Maximization: $\boldsymbol{\theta}^{(t)} = \arg \max_{\boldsymbol{\theta}} \mathcal{L}\left(\boldsymbol{\theta}, \boldsymbol{\phi}^{(t)} ; \mathbf{x}^k\right)$.}
\end{enumerate}
In the expectation step a differentiable transform $g_{\phi}(\mathbf{x}^k, \boldsymbol{\epsilon}^{j, k})$ of auxiliary samples $\boldsymbol{\epsilon}^{j, k} \sim p(\boldsymbol{\epsilon})$ is used to estimate the expectation over the inference network distribution. Coined the ``reparameterization trick,'' this technique has been show to reduce the variance of the gradient estimates through the inference network \cite{kingma_auto-encoding_2014, rezende_stochastic_2014}. In practice, the two steps of the EM algorithm are often approximated concurrently with a gradient step using a single sample $J = 1$.

In the next sections, we will describe 1) a method for reparameterization that results in sparse latent features and 2) a new sampling procedure in the expectation step that improves performance for distributions with sparse priors.

\subsection{Reparameterization for Thresholded Samples}
\begin{figure}[t!]
    \centering
    \includegraphics[scale=0.43]{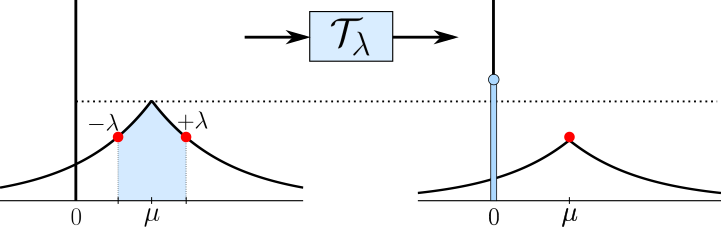}
    %\vspace{-0.6cm}
    \caption{Visual depiction of the shifted soft-threshold applied to a Laplacian distribution. Probability mass $\mu \pm \lambda$ is collapsed to the origin. The red points in the left are shifted to meet at $\mu$ on the right.}
    \label{fig:threshold_fig}
\end{figure}
\label{sec:inference}
A favorable choice of sparse prior distribution is the Spike-and-Slab $p(z_i) = \gamma p(s_i) + (1 - \gamma)\delta(z_i)$ \cite{mitchell_bayesian_1988}, which allows sparse random variables to be sampled by first sampling a Bernoulli random variable with probability $\gamma$. If the random variable is non-zero, then a sample is drawn from the slab distribution $p(s_i)$. A Spike-and-Slab is favorable to setting $p(z_i)$ equal to Normal, Laplacian, or Cauchy distributions because these latter distributions will never result in samples with features exactly equal to zero. This allows one to benefit from precise sparsity for downstream computations (e.g., decoding, image generation). Unfortunately, current BBVI approaches depend on continuous approximations (controlled by temperature parameter $\tau$) to each Bernoulli random variable $\tau$ \cite{tonolini_variational_2020, jang_categorical_2017, maddison_concrete_2017}. Tuning $\tau$ requires a trade-off between variance and bias when estimating gradients, potentially leading to poor performance.

Rather than parameterize $p(z_i)$ with discrete random variables, we propose an approach that applies shifted soft-thresholding of samples drawn from a Laplacian distribution. Let $p(s) = \mbox{Laplace}(\mu, b)$ be a Laplacian pdf with shift $\mu$ and scale $b$. We can draw a sample $s \sim p(s)$ and apply the shifted soft-threshold function:
\begin{align}
\label{eq:shift_soft_threshold}
\mathcal{T}(s ; \lambda, \mu) = & \; \mbox{sign}(s - \mu)\max \bigl(|s - \mu| - \lambda, 0 \bigr) \\
& + \mathbb{I}\left[|s - \mu| > \lambda \right]\mu. \nonumber
\end{align}
For ease of notation, we will use the shorthand $\mathcal{T}_\lambda(s)$ to refer to $\mathcal{T}(s; \lambda, \mu)$. Intuitively, this function is a soft-threshold around $\mu$, where points mapped to $\mu$ are set exactly to zero. This is visualized in Figure~\ref{fig:threshold_fig}. When $\mu=0$, this function is precisely the soft-threshold function \cite{donoho_ideal_1994}. Next, we will show how this shifted soft-threshold can be used to reparameterize to a Spike-and-Slab without needing discrete random variables.

\newtheorem{prop}{Proposition}[section]

\begin{prop}
Samples from a Laplace distribution $s \sim p(s)$ passed through a shifted soft-threshold $z$ = $\mathcal{T}_{\lambda}(s)$ are equivalently distributed as a Spike-and-Slab distribution $z \stackrel{i.i.d.}{\sim} p(z)$.
\end{prop}
Note that under this rule for $z$, $p(z \mid |s - \mu | \leq \lambda) = \delta(z)$, where the spike probability $(1 - \gamma)$ can be found by computing $p(|s - \mu | \leq \lambda)$:
\begin{align}
p(|s - \mu | \leq \lambda) & = 2\int_{0}^{\lambda} \frac{1}{2b} \exp\left(-s b^{-1}\right) ds \nonumber \\
& = 1 - \exp\left(-\lambda b^{-1}\right) = (1 - \gamma).
\label{eq:spike_prob}
\end{align}
We show in Appendix~\ref{sec:appendix_threshold_der} that $p(z \mid | s - \mu | > \lambda)$ is distributed as $p(s)$. Combining these facts and marginalizing over $p(s)$ yields:
\begin{equation}
z \sim  \frac{\gamma}{2b}\exp\left(\frac{- | z - \mu| }{b}\right) + (1 - \gamma) \delta(z). \nonumber
\end{equation}

Using this technique, we can sample sparse random variables $\mathbf{z}^k$ for a corresponding input data $\mathbf{x}^k$ from an inference network trained via the EM algorithm~(\ref{eq:elbo}). For a given input data sample, we encode the parameters of a base distribution that is either a Gaussian or Laplacian via $q_{\boldsymbol{\phi}}(\mathbf{s} \mid \mathbf{x}^k)$. These can be used to both compute the KL divergence and to reparameterize a sample to be passed through the shifted soft-threshold. The threshold parameter can be either set as a fixed hyperparameter $\boldsymbol{\lambda} = \boldsymbol{\lambda}_0$ (equating to a fixed prior on the spike probability) or learned via variational inference with a factorial gamma prior $p(\boldsymbol{\lambda}) = \prod_{i=1}^d \Gamma(\alpha_0, \frac{\alpha_0}{\lambda_0})$ \cite{jankowiak_pathwise_2018,garrigues_group_2010}. In the case where we perform inference on $\boldsymbol{\lambda}$, we assume independence with the base distribution $p(\mathbf{s})$.

One challenge during training is dealing with the non-differentiable $T_{\boldsymbol{\lambda}}$ in the first term of~(\ref{eq:elbo}). We have found that the subgradient of this function leads to shrinkage of the base distribution parameters, eventually leading to numerical instability or posterior collapse \cite{lucas_dont_2019}. For further discussion on this issue, refer to Appendix~\ref{sec:appendix_subg}. As a result, we employ a straight-through estimator \cite{bengio_estimating_2013, jang_categorical_2017, oord_neural_2018}, skipping the shifted soft-threshold when we differentiate the loss by passing the gradient from the generator directly to the base distribution parameters. This is depicted in Figure~\ref{fig:intro_fig}.

This results in our final training objective with a reweighted KL term \cite{higgins_vae_2017}. Following the notation of \cite{oord_neural_2018}, we denote by $\mbox{sg}[\cdot]$ the stopgradient operator (identity in the forward pass with partial derivatives equal to zero).
\begin{equation}
\label{eq:ste_threshold}
\widetilde{\mathbf{z}}^k = \mathbf{s}^k + \mathcal{T}_{\boldsymbol{\lambda}^k}\left(\mbox{sg}\left[\mathbf{s}^k\right]\right) - \mbox{sg}\left[\mathbf{s}^k \right]
\end{equation}
\begin{align}
\label{eq:elbo_train}
\mathcal{L}\left(\boldsymbol{\theta}, \boldsymbol{\phi} ; \mathbf{x}^k \right) = ~ & \mathbb{E}_{q_{\boldsymbol{\phi}}}\left[\log p_{\boldsymbol{\theta}}(\mathbf{x}^k \mid \widetilde{\mathbf{z}}^k)\right] \nonumber \\
& - \beta_1 D_{KL}\bigl(q_{\phi}(\mathbf{s} \mid \mathbf{x}^k) ~||~ p(\mathbf{s})\bigr)  \\
& - \beta_2 D_{KL}\bigl(q_{\phi}(\boldsymbol{\lambda} \mid \mathbf{x}^k) ~||~ p(\boldsymbol{\lambda})\bigr) \nonumber
\end{align}
Note that in the case where we fix $\boldsymbol{\lambda}^k = \boldsymbol{\lambda}_0$, the second term in~(\ref{eq:ste_threshold}) has no gradient and the third term in~(\ref{eq:elbo_train}) is omitted. More details on reparameterization for the sampling procedure $(\mathbf{s}^k, \boldsymbol{\lambda}^k) \sim q_{\boldsymbol{\phi}}(\mathbf{s}, \boldsymbol{\lambda} \mid \mathbf{x}^k)$ and on computing the loss terms in~(\ref{eq:elbo_train}) are included in Appendix~\ref{sec:appendix_reparam}.

\begin{algorithm}[tb]
 \caption{Training with Thresholded Samples}
 \label{alg:training_alg}
\begin{algorithmic}
 \STATE {\bfseries Input:} Training batch $\mathbf{x}^k$, threshold hyper-parameter $\boldsymbol{\lambda}_0$, whether to use a GammaPrior (along with gamma hyper-prior $\boldsymbol{\alpha}_0$), network initializations $(\boldsymbol{\phi}^0, \boldsymbol{\theta}^0)$, number of samples $J$, and number of iterations $T$.
 \FOR{$t=1$ {\bfseries to} $T$}
  \FOR{$j=1$ {\bfseries to} $J$}
    \STATE $\boldsymbol{\epsilon}^{j, k} \sim p(\boldsymbol{\epsilon})$
   \IF{GammaPrior}
   \vspace{0.1cm}
    \STATE $\left(\mathbf{s}^{j, k}, \boldsymbol{\lambda}^{j, k} \right) \gets g_{\boldsymbol{\phi}}\left(\mathbf{x}^k, \boldsymbol{\epsilon}^{j, k}\right)$
   \ELSE
     \STATE $\mathbf{s}^{j, k} \gets g_{\boldsymbol{\phi}}\left(\mathbf{x}^k, \boldsymbol{\epsilon}^{j, k}\right)$
     \STATE $\boldsymbol{\lambda}^{j, k} \gets \boldsymbol{\lambda}_0$
   \ENDIF
   \STATE $\mathbf{\widetilde{z}}^{j, k} \gets \mathbf{s}^{j, k} + \mathcal{T}_{\boldsymbol{\lambda}^{j, k}}\left(\mbox{sg}\left[\mathbf{s}^{j, k}\right]\right) - \mbox{sg}\left[\mathbf{s}^{j, k} \right]$
   \STATE $\mathcal{L}^{j, k} \gets \log p_{\boldsymbol{\theta}}(\mathbf{x}^k \mid \widetilde{\mathbf{z}}^{j,k}) - \beta D_{KL}$
  \ENDFOR
  \STATE $\widehat{\mathcal{L}}^k = \arg\max_j \mathcal{L}^{j, k}$
  \STATE $(\boldsymbol{\phi}^t, \boldsymbol{\theta}^t) = \arg\max \widehat{\mathcal{L}}^k$
\ENDFOR
\end{algorithmic}
\end{algorithm}

\subsection{Max ELBO Sampling}
\label{sec:sample}
A favorable property of MAP estimates is that priority of which latent features are non-zero is given to those that best represent features in input data. These developed features lead to the highest decrease in loss, a property exploited in greedy MAP inference procedures \cite{tropp_signal_2007}. Unfortunately, BBVI does not inherently have this property, with the factorial prior in the KL divergence term in~(\ref{eq:elbo_train}) encouraging all features to be equally likely for each input data sample. To address this, we propose a sampling procedure that biases towards reuse of developed features.

Before introducing our approach, we note the standard approach in BBVI for using multiple samples to approximate the expectation with respect to $q_{\boldsymbol{\phi}}$ in~(\ref{eq:elbo_train}). Given a sampling budget $J$, one may draw $J$ i.i.d. samples $\mathbf{z}^{j, k} \sim q_{\boldsymbol{\phi}}(\mathbf{z} \mid \mathbf{x}^k)$ from the posterior and compute the ELBO for each sample.
\begin{equation}
\mathcal{L}^{j, k} = \log p_{\boldsymbol{\theta}}(\mathbf{x}^{j, k} \mid \mathbf{z}^{j, k}) - \beta D_{KL}\bigl(q_{\phi}(\mathbf{z}^{j, k} \mid \mathbf{x}^k) ~||~ p(\mathbf{z})\bigr)
\end{equation}
Then, to estimate the training objective, one may average over the loss from each sample:
\begin{equation}
\widehat{\mathcal{L}}^k_{\tiny{avg}} = \frac{1}{J}\sum_{j=1}^J \mathcal{L}^{j, k}.
\label{eq:avg_sample}
\end{equation}
In the case where we have a sparse prior, each of the $J$ samples taken here will have different support and thus encourage development of different latent features.

Rather than give each sample equal probability, we introduce a new sampling strategy motivated by the approximation to expectations utilized in \cite{olshausen_emergence_1996, connor_variational_2020}. In these works, it is observed that the selected prior distribution concentrates most of its probability mass around the maximum value. This leads to an approximation of the expectation over the prior latent variable with a delta at the max value $\mathbf{\widehat{z}}$, motivating a MAP estimate:
\begin{align}
\log \mathbb{E}_{p(\mathbf{z})}\left[p_{\boldsymbol{\theta}}(\mathbf{x} \mid \mathbf{z}) \right] & \approx \log \int_{z}p_{\boldsymbol{\theta}}(\mathbf{x} \mid \mathbf{z})p(\mathbf{z})\delta(\widehat{\mathbf{z}})d\mathbf{z} \nonumber \\
& = \max_{\mathbf{z}} \log p_{\boldsymbol{\theta}}(\mathbf{x} \mid \mathbf{z}) + \log p(\mathbf{z}).
\end{align}

Although our training setting differs in that we take our expectation over the variational posterior $\mathbb{E}_{q_{\boldsymbol{\phi}}}$, in many cases we explicitly regularize this distribution with a KL divergence penalty against a sparse prior. To this same end, rather than approximate our expectation $\mathbb{E}_{q_{\boldsymbol{\phi}}}$ with the average over several samples, we approximate it with a single sample with the highest likelihood:
\begin{equation}
\widehat{\mathcal{L}}_{\tiny{max}}^k = \max_j \mathcal{L}^{j, k}.
\label{eq:max_sample}
\end{equation}

Although this provides a biased estimate of the loss, we will show in later sections that this heuristic provides an increase in performance and latent feature reuse during training. We combine this method with our sampling approach to outline our full training procedure in Algorithm~\ref{alg:training_alg}.

\section{Experiments}
\subsection{Linear Generator}

\subsubsection{Sparse Coding Performance}

\begin{table*}[t!]
\centering
\caption{\label{linear_gen_comp}Performance of different inference methods with a linear generator on whitened image patches. Validation loss computed on objective~(\ref{eq:fista_baseline}). Multi-information measures the statistical efficiency of inferred codes. The importance weighted autoencoder (IWAE) loss \cite{burda_importance_2016} is a tight bound on the ELBO, but does not necessarily depict success in the sparse coding task. A table with standard deviations for $J=20$ is included in Table~\ref{tab:quant_linear_std} in Appendix~\ref{sec:extra_linear_results}}
\begin{tabular}{@{}c|cc|cc|cc@{}}
\toprule
Method/Prior   Distribution & \multicolumn{2}{c|}{Validation Loss}   & \multicolumn{2}{c|}{Multi-Information} & \multicolumn{2}{c}{IWAE Loss}                    \\
& J=1 & J=20 & J=1 & J=20 & J=1 & J=20                          \\
\midrule
FISTA (baseline)            & 1.01E+02 & --        & 8.75E+01 & --        & --        & --   \\
Gaussian                    & 1.35E+03 & 1.35E+03  & 7.34E+02 & 7.36E+02  & 2.57E-01  & 2.16E-01  \\
Laplacian                   & 5.96E+02 & 5.79E+02  & 5.36E+02 & 5.34E+02  & \textbf{2.20E-01}  & 2.22E-01  \\
Spike-and-slab              & 2.52E+02 & 2.39E+02  & 1.94E+02 & 1.96E+02  & 2.41E-01 & \textbf{2.12E-01}  \\
Thresholded Gaussian        & 2.32E+02 & 2.30E+02  & \textbf{1.67E+02} & \textbf{1.76E+02}  & 1.52E+00  & 1.33E+00  \\
Thresholded Gaussian+Gamma  & 2.85E+02 & 2.70E+02  & 2.35E+02 & 2.29E+02  & 1.17E+00  & 1.13E+00  \\
Thresholded Laplacian       & \textbf{1.98E+02} & \textbf{1.94E+02}  & 1.80E+02 & 1.91E+02 & 9.30E-01  & 1.13E+00  \\
Thresholded Laplacian+Gamma & 2.23E+02 & 2.11E+02  & 2.38E+02 & 2.33E+02  & 9.81E-01  & 1.12E+00  \\    \bottomrule
\end{tabular}
\end{table*}

In our first experiment, we test different inference methods on sparse coding of the whitened image patches used in \cite{olshausen_emergence_1996}. We train on 80,000 16x16 training patches with a dictionary of 256 elements (i.e., a linear generator $\boldsymbol{\theta} = \mathbf{A} \in \mathbb{R}^{256 \times 256}$). We start in this simplified setting since it allows us to compare against baseline sparse coding strategies (that use MAP estimates) and so we can analyze the dictionary we learn for each method. As a baseline, we infer coefficients $\mathbf{\widehat{z}}^k$ using the FISTA algorithm \cite{beck_fast_2009} to optimize objective~(\ref{eq:fista_baseline}). This objective includes a data fidelity term, sparsity-encouraging $\ell_1$ penalty on latent features, and a Frobenius norm regularizer on the dictionary to prevent unbounded growth:
\begin{equation}
\label{eq:fista_baseline}
\min_{\mathbf{A}} \Vert \mathbf{x}^k - \mathbf{A}\mathbf{\widehat{z}}^k\Vert_2^2 + \lambda \Vert \widehat{\mathbf{z}}^k \Vert_1 + \kappa \Vert \mathbf{A}\Vert_F^2.
\end{equation}
In all of our experiments, we set the sparsity hyperparameter such that roughly $10\%$ features are non-zero for each batch of data. For FISTA, this occurs when $\lambda = 20$. We use this objective to measure validation performance of our variational methods.

To train our inference networks, we use an ELBO objective with a Frobenius norm regularizer on the dictionary:
\begin{equation}
\min_{\mathbf{A}, \boldsymbol{\phi}}~ -\mathcal{L}(\mathbf{A}, \boldsymbol{\phi}; \mathbf{x}^k) + \kappa ||\mathbf{A}||_F^2
\end{equation}
\begin{equation}
\log p_{\mathbf{A}}(\mathbf{x}^k | \mathbf{z}^k) = -\Vert \mathbf{x}^k - \mathbf{A} \mathbf{z}^k \Vert_2^2,
\end{equation}
where $\mathbf{z}^k$ is either found via the inference procedure outlined in section~\ref{sec:inference} or by previous methods that used Gaussian \cite{kingma_auto-encoding_2014}, Laplacian \cite{barello_sparse-coding_2018}, or Spike-and-Slab \cite{tonolini_variational_2020} prior distributions. For the Spike-and-Slab prior, we approximate each Bernoulli random variable with a straight-through estimator of a Gumbel-Softmax sample \cite{jang_categorical_2017}. For each method, we apply the sampling procedure outlined in Section~\ref{sec:sample} with $J = 1$ and $J = 20$. Details on training hyper-parameters are provided in Appendix~\ref{sec:linear_train_detail}.

%For FISTA, we set $\kappa = 1\mathrm{E}{-03}$. For the variational methods, we set $\kappa = 1\mathrm{E}{-04}$, and the scale parameter of our prior $p(\mathbf{z})$ equal to $0.1$ when applicable.
\begin{figure}[h]
\centering
    \subfigure[FISTA]{\includegraphics[width=0.15\textwidth]{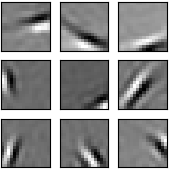}}
    ~
    \subfigure[Gaussian]{\includegraphics[width=0.15\textwidth]{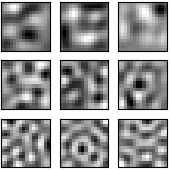}}
    ~
    \subfigure[Laplacian]{\includegraphics[width=0.15\textwidth]{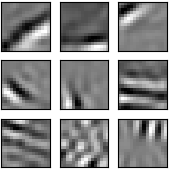}}

    \subfigure[Spike-and-Slab]{\includegraphics[width=0.15\textwidth]{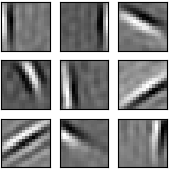}}
    ~
    \subfigure[Thresholded Gaussian]{\includegraphics[width=0.15\textwidth]{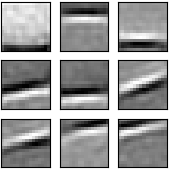}}
    ~
    \subfigure[Thresholded Laplacian]{\includegraphics[width=0.15\textwidth]{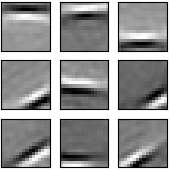}}
    %\vspace{-0.5cm}
    \caption{Entries of the learned linear generator (dictionary) for different approaches for inferring sparse codes. FISTA uses an iterative MAP estimate that optimizes objective~(\ref{eq:fista_baseline}) directly. Others use a variational approach with different prior distributions using $J=20$ samples. The Spike-and-Slab, Thresholded Gaussian, and Thresholded Laplacian methods learn dictionaries with most entries visually resembling the dictionary learned by FISTA.}\label{fig:learn_dict}
    %\vspace{-0.3cm}
\end{figure}

Figure~\ref{fig:learn_dict} plots the columns of the learned dictionary for each inference method for $J=20$. As expected, FISTA learns dictionary entries that qualitatively resemble Gabor wavelets \cite{olshausen_emergence_1996}. The Laplacian prior \cite{barello_sparse-coding_2018} learns a few dictionary entries that resemble wavelets, with most entries resembling noise. Although the entries learned by all sparse priors resemble wavelets, only the Thresholded Laplacian is capable of learning this structure with $J=1$. Full learned dictionaries for all methods under both sampling budgets are visualized in Appendix~\ref{sec:extra_linear_results}.

Table~\ref{linear_gen_comp} compares average quantitative performance over three training runs for each inference method. The Laplacian prior with a fixed threshold performs the best among variational methods on objective  (\ref{eq:fista_baseline}). We note that methods that learn the threshold parameters tend to increase the number of active latent features, leading to an increase in L1 penalty in the validation loss. Although the Spike-and-Slab prior minimizes the importance weighted (IWAE) loss from \cite{burda_importance_2016} with $K=200$, it performs worse on the sparse coding objective. We note that the IWAE loss measures the tightness of the ELBO bound (reconstruction + KL loss), whereas the thresholding in our method has a trade-off between deviation of the KL loss and posterior expressivity.

To measure the statistical efficiency in the inferred codes, we follow previous work in estimating the multi-information, $\sum_{i=1}^{d} h(z_i) - h(\mathbf{z})$, of the inferred codes \cite{fallah_learning_2020, eichhorn_natural_2009, bethge_factorial_2006}. This quantity is minimized when the features have high joint entropy while being independent of one another, denoting a reduction in redundancy. Among variational methods, the Thresholded Gaussian minimizes this quantity.

To further understand the benefit of our proposed method, we investigate the quality of the gradient estimates through both the dictionary and inference network at the end of training. To measure this, we use the signal-to-noise ratio (SNR) metric proposed in \cite{rainforth_tighter_2019}:
\begin{equation*}
\mbox{SNR}(\theta) = \left| \mathbb{E}\left[\nabla_{\theta} \mathcal{L}\right] / \sigma\left[\nabla_{\theta} \mathcal{L}\right] \right|.
\end{equation*}
Where $\theta$ is a single parameter from our dictionary or inference network and $\mathcal{L}$ is our training loss. In our experiments, we average the SNR over all parameters in our inference network or dictionary.

\begin{figure}[b!]
\centering
    \subfigure[Inference Network Gradient SNR]{\hspace{-0.4cm}\includegraphics[width=0.5\textwidth]{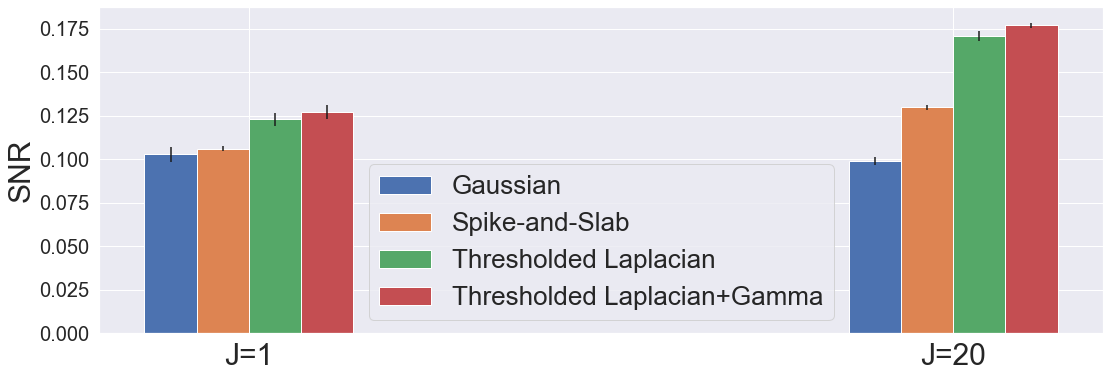}}
    %\vspace{-0.2cm}
    \subfigure[Generator Gradient SNR]{\hspace{-0.5cm}\includegraphics[width=0.5\textwidth]{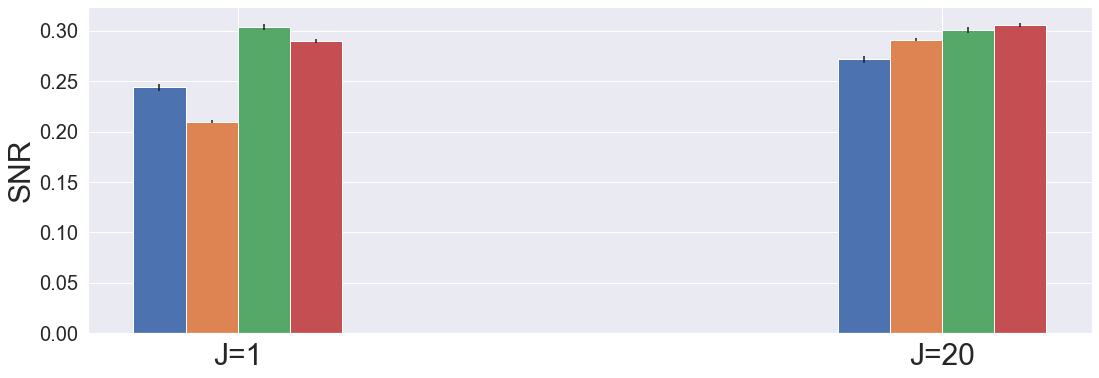}}
    %\hspace{-0.5cm}
    \caption{Comparison of the SNR of the (a) inference network and (b) linear generator gradients at the end of training. Thresholded methods perform the best, with all sparse priors experiencing higher SNR with $J=20$ samples under max ELBO sampling.}\label{fig:grad_snr}
\end{figure}

The benefit of measuring the gradient quality through a relative metric like SNR is that it weights the variance by the importance of each parameter. For example, dictionary entries that are rarely activated for a given input data will have a low expected gradient, but may have high variance. We evaluate the SNR over $1000$ sparse code samples for each data point in our validation set. Figure~\ref{fig:grad_snr} compares the SNR for both the dictionary and inference network for various methods. It can be seen that the Spike-and-Slab suffers from low SNR in the case of $J=1$, with a significant increase for $J=20$. We speculate that this metric explains the lack of clear wavelets in the dictionary learned by the Spike-and-Slab for $J=1$. %Finally, we include a comparison of the thresholded Laplacian with a subgradient estimator (where we estimate the gradient through the shifted soft-threshold) and see a dramatic decrease in the SNR.

\subsubsection{Sampling Performance}
\begin{figure}[]
\centering
    \subfigure[J=20]{\includegraphics[width=0.45\textwidth]{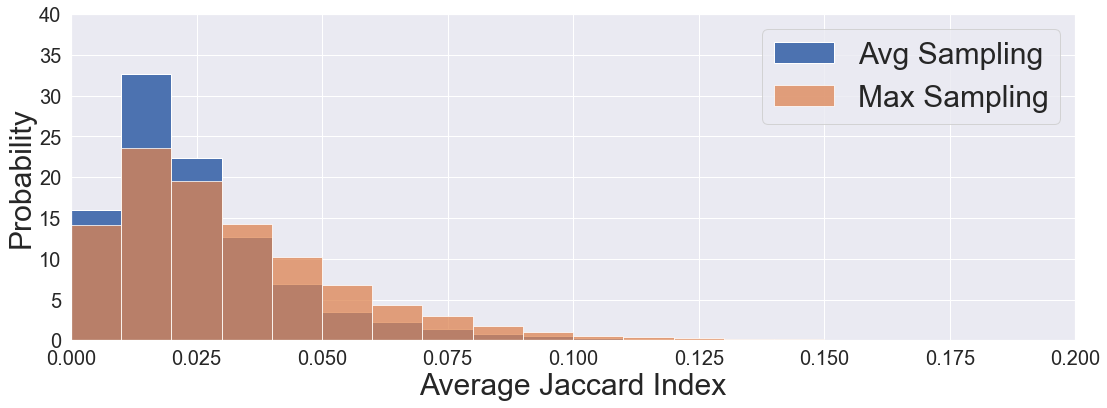}}
    %\hspace{-0.5cm}
    \subfigure[J=100]{\includegraphics[width=0.45\textwidth]{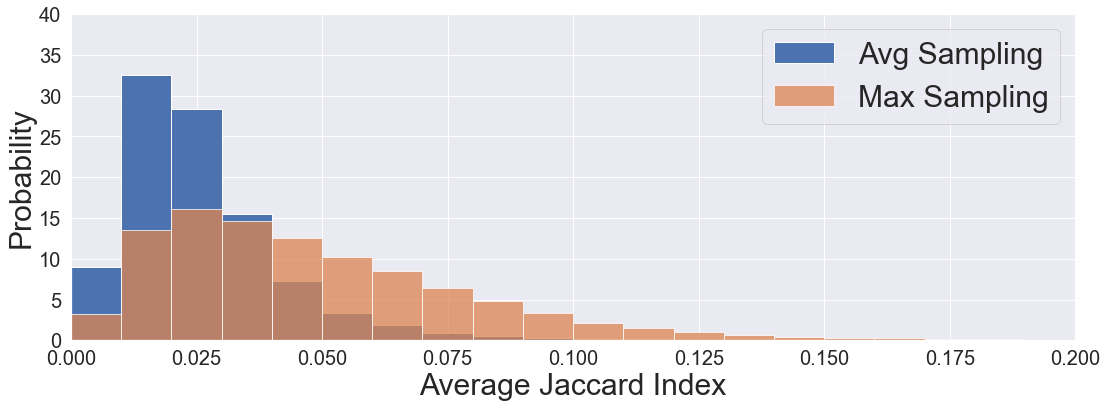}}
    %\vspace{-0.4cm}
    \caption{Histogram of the average pairwise Jaccard index over 20 forward passes on two trained inference networks. The pairwise Jaccard index measures the consistency of the support set of inferred codes. Max ELBO sampling leads to a higher average Jaccard index, indicating reuse of developed latent features.}\label{fig:pairwise_jaccard}
    %\vspace{-0.3cm}
\end{figure}

To test whether our sampling procedure leads to more feature reuse, we measure the consistency in the features used over several forward passes for fixed input data. Each forward pass leads to a sparse code with a different support set, indicating use of different features. Hence, we can compute the Jaccard index between the support set of several samples $\mathbf{z}^{1,k}, \dots, \mathbf{z}^{J,k} \sim q_{\phi}(\mathbf{z} \mid \mathbf{x}^k)$ for fixed input to measure consistency \cite{jaccard_distribution_1912}. Intuitively, the Jaccard index measures similarity between discrete sets, providing a value in $[0, 1]$ based on how similar the support is over different forward passes. Let $S(\mathbf{z}^{j,k})$ denote the support of $\mathbf{z}^{j,k}$ (i.e. $i \in S(\mathbf{z}^{j,k})$ if $z_i^{j,k} \neq 0$), the Jaccard index is computed as:
\begin{equation}
C(\mathbf{z}^{m,k}, \mathbf{z}^{n,k}) = \frac{\left|S(\mathbf{z}^{m,k}) \cap S(\mathbf{z}^{n,k})\right|}{\left|S(\mathbf{z}^{m,k}) \cup S(\mathbf{z}^{n,k})\right|}.
\end{equation}
For any given input data $\mathbf{x}^k$ we can draw $J$ samples from our inference network and compute the average pairwise Jaccard index as $\frac{1}{J(J-1)}\sum^J_{m}\sum^J_{n\neq m} C(\mathbf{z}^{m,k}, \mathbf{z}^{n,k})$. To normalize for the fact that each training methods leads to a different number of non-zero dictionary entries (e.g., an inference network that only uses a subset of the available support is more likely to be consistent), we measure consistency using the support of features with $\Vert \mathbf{a}_i \Vert_2^2 > 1\mathrm{E}\mathrm{-01}$.

We use this metric to compare the consistency of inference networks trained under the average ELBO scheme from~(\ref{eq:avg_sample}) and the proposed max ELBO scheme in~(\ref{eq:max_sample}) using $J=[20, 100]$ samples. We observe that both methods result in $10\%$ of features non-zero on average.

We plot the consistency on the validation set averaged over three training runs as a histogram in Figure~\ref{fig:pairwise_jaccard}. It can be seen that the max ELBO sampling procedure leads to higher consistency (indicating reuse of developed features) compared to average sampling. Qualitatively, the dictionary entries learned in the average sampling scheme has several repeated entries, indicating poor reuse in comparison to the qualitatively diverse entries learned by max sampling. We visualize the full dictionary for both in Figure~\ref{fig:full_sample_dict} in Appendix~\ref{sec:extra_linear_results}.
\subsection{DNN Generator}

\begin{figure}[t!]
\centering
    \subfigure[Gaussian]{\includegraphics[width=0.45\textwidth]{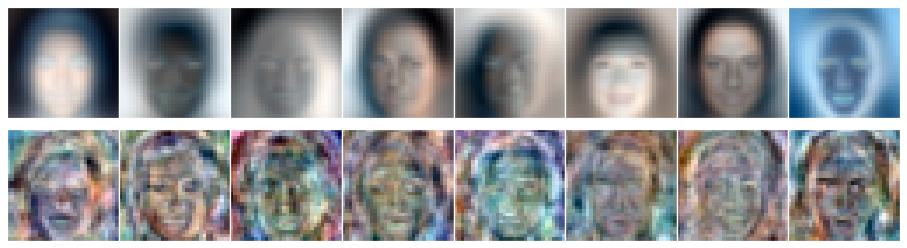}}
    \subfigure[Thresholded Gaussian+Gamma]{\includegraphics[width=0.45\textwidth]{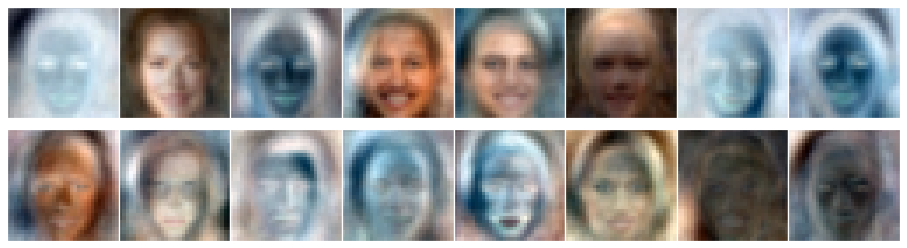}}
    \caption{Selected entries of an estimated dictionary for variational autoencoders trained with a (a) Gaussian prior and (b) Thresholded Gaussian+Gamma prior with a $512$ dimensional latent space\footnotemark. For each method, the first row denotes dictionary entries with the highest magnitude and the bottom row denotes random entries from the middle of the dictionary.}\label{fig:celeb_est_dict}
    %\vspace{-0.3cm}
\end{figure}
\footnotetext{This visualization is from a training run where the sparsity prior is set to encourage 5\% features non-zero.}

We next investigate the efficacy of variational sparse coding in unsupervised learning on the Fashion MNIST (FMNIST) \cite{https://doi.org/10.48550/arxiv.1708.07747} and CelebA datasets \cite{liu2015faceattributes}. We follow a similar methodology as the previous section, except that we apply a DNN generator in place of the linear dictionary (more details in Appendix~\ref{sec:dnn_train_detail}). Following the principle of overcomplete representations in sparse coding, where the robustness of latent features increases with the dimensionality \cite{olshausen_highly_2013}, we compare performance as we sweep dimensionality.

\begin{figure*}[hbt!]
\centering
    \subfigure[FMNIST MI]{\includegraphics[width=0.24\textwidth]{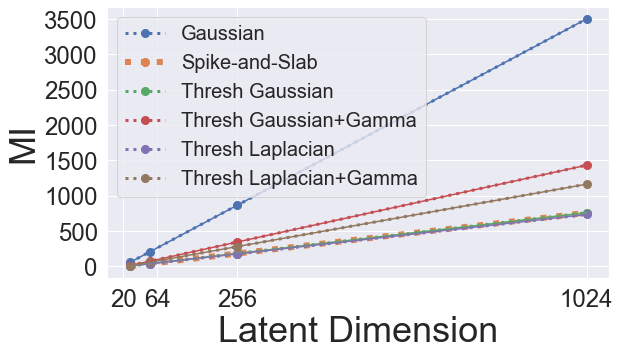}}
    \subfigure[FMNIST FID]{\includegraphics[width=0.24\textwidth]{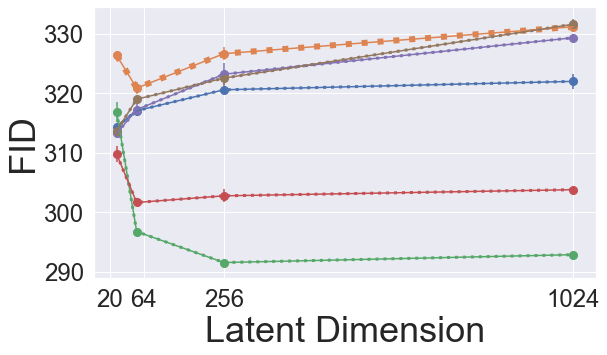}}
    \subfigure[FMNIST MSE]{\includegraphics[width=0.24\textwidth]{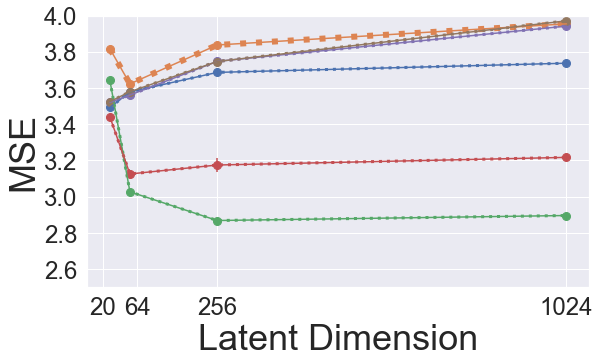}}
    \subfigure[FMNIST MSE per \% Sparsity]{\includegraphics[width=0.24\textwidth]{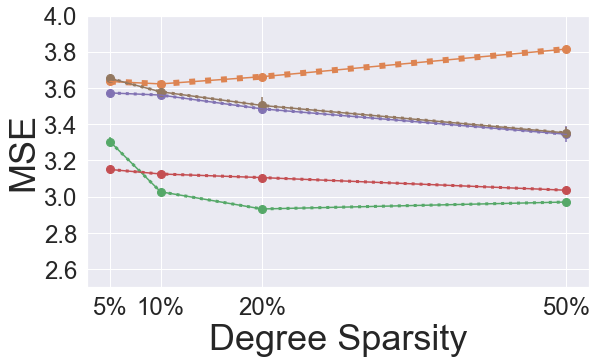}} \\

    \subfigure[CelebA MI]{\includegraphics[width=0.24\textwidth]{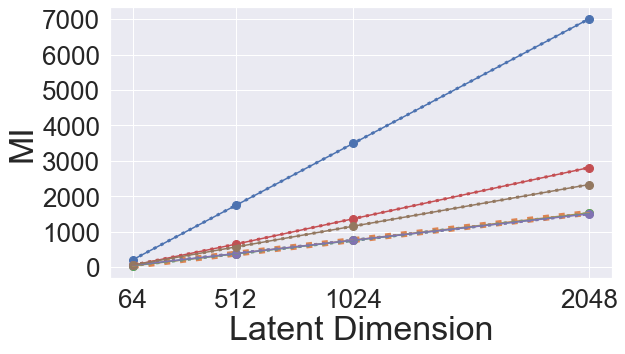}}
    \subfigure[CelebA FID]{\includegraphics[width=0.24\textwidth]{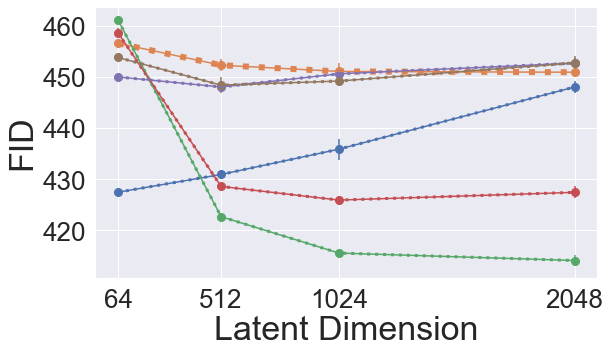}}
    \subfigure[CelebA MSE]{\includegraphics[width=0.24\textwidth]{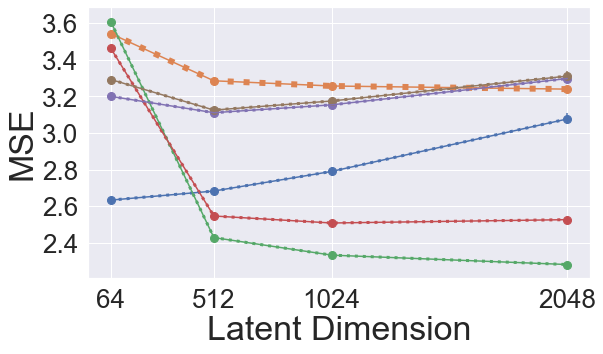}}
    \subfigure[CelebA MSE per \% Sparsity]{\includegraphics[width=0.24\textwidth]{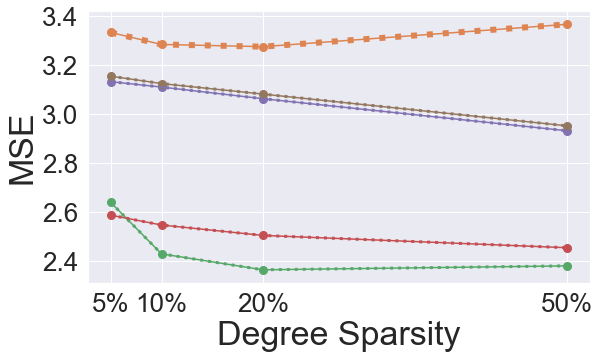}}

    \caption{Comparison of variational autoencoder with different prior distributions as the latent feature dimensionality increases on the Fashion MNIST (a-d) and CelebA (e-h) datasets. (a, e) Multi-information scales at a higher rate with the Gaussian prior, indicating inferior statistical efficiency at higher latent dimensions. (b, f) The Fr\'echet Inception Distance of images reconstructed by learned DNN generator. (c, g) Mean squared error on validation data for a fixed degree of sparsity. (d, h) Mean squared error for a fixed dimensionality with an increase in the degree of sparsity (indicating the \% of non-zero latent features set by the prior).}\label{fig:celeba_quant}
\end{figure*}

Figure~\ref{fig:celeba_quant} shows the quantitative performance of different prior methods trained with either increasing latent dimensionality or percentage of latent features non-zero. For both datasets, as the dimensionality increases with a fixed degree of sparsity, it can be seen that the Gaussian prior (corresponding to a variational autoencoder \cite{kingma_auto-encoding_2014, higgins_vae_2017}) has a significant increase in redundancy, measured via the multi-information, with a slight increase in the validation Fr\'echet inception distance (FID) \cite{heusel_gans_2017} and MSE. Alternatively, all the sparse priors show superior scaling with dimensionality, with the Thresholded Gaussian and Thresholded Gaussian+Gamma depicting superior performance to the Spike-and-Slab model from \citet{tonolini_variational_2020}. Additionally, all thresholded methods scale better with MSE as the degree of sparsity (corresponding to the \% of features non-zero) increases.

We observe the same qualitative generative interpretability when sweeping each individual latent feature as in previous work that investigates Gaussian priors \cite{higgins_vae_2017} and sparse priors \cite{tonolini_variational_2020}. In these works, it is observed that each latent feature corresponds to a specific semantic change in the generated image (such as changes in skin color, background color, hairstyle, etc.). Unfortunately, in the case where one trains a DNN generator, they are unable to observe the generator features the same way one could with the columns of the linear dictionary.

To this end, we estimate a linear dictionary for each fully trained CelebA inference network to visualize what is encoded by each latent feature. Given a collection of data samples $\mathbf{X} = \left [\mathbf{x}^1, \dots, \mathbf{x}^n \right]^T$ and corresponding inferred codes $\mathbf{Z} = \left [\mathbf{z}^1, \dots, \mathbf{z}^n \right]^T$ from an inference network with fixed weights fully trained with a DNN generator $\mathbf{z}^k \sim q_{\boldsymbol{\phi}}(\mathbf{z} \mid \mathbf{x}^k)$, one can estimate a linear dictionary as \cite{isely_deciphering_2010, fallah_learning_2020}:
\begin{equation*}
\mathbf{X} = \mathbf{A} \mathbf{Z}
\end{equation*}
\begin{equation*}
\widehat{\mathbf{A}} = \left(\mathbf{X} \mathbf{Z}^T\right) \left(\mathbf{Z}\mathbf{Z}^T \right)^{-1}.
\end{equation*}

We use this technique with each trained inference network to analyze which features in the training dataset are represented by each latent feature. We plot selected entries from dictionaries estimated for the Gaussian and Thresholded Gaussian+Gamma in Figure~\ref{fig:celeb_est_dict}. Interestingly, the first eight entries from the Gaussian prior resemble the qualitative features observed in the $\beta$-VAE \cite{higgins_vae_2017}. Poor statistical efficiency in the Gaussian prior indicates that certain latent features are not encoding useful information from training data, leading to corresponding estimated dictionary entries that resemble noise (visualized in the second row). Meanwhile, all the entries for the sparse prior more clearly resemble features in the training dataset. While the qualitative results here are similar to previous research comparing learned features from principal component analysis (PCA) and independent component analysis (ICA) on face datasets \cite{bartlett_face_2002}, a detailed quantitative comparison with those results is beyond the scope of the present paper. We visualize more estimated dictionary plots in Appendix~\ref{sec:dnn_extra_results}.

\begin{figure}[]
\centering
    \includegraphics[width=0.48\textwidth]{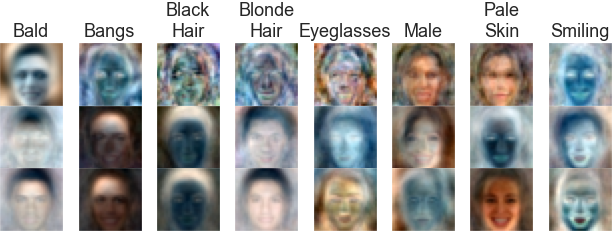}
    %\vspace{-0.3cm}
    \caption{Estimated dictionary entries for latent features with the highest absolute point-biserial correlation with the CelebA binary attribute labels. From top to bottom, the rows denote the Gaussian prior, Thresholded Gaussian prior, and Thresholded Gaussian+Gamma prior.}\label{fig:celeba_dict_corr}
    %\vspace{-0.3cm}
\end{figure}

Finally, we use the binary attribute labels included with the CelebA dataset to measure correlation between latent features and attributes in the input data. Given vectors of $40$ binary attributes for each data sample $\mathbf{y}^1, \dots, \mathbf{y}^N$ and corresponding latent features $\mathbf{z}^1, \dots, \mathbf{z}^N$ from a trained inference network, we can compute the point-biserial correlation coefficient between each attribute $m$ and each dimension of latent feature $n$ as:
\begin{equation*}
\rho^{m,n}_{pb} = \frac{M_1 - M_0}{s_N}\frac{\sqrt{n_1 n_0}}{\sqrt{N}}.
\end{equation*}
Where $M_1$ and $M_0$ are the sample mean of latent features $z^{k}_n$ corresponding to when attribute label $y^k_m$ equals one or zero, respectively. $s_N$ is the sample standard deviation over all features, and $n_1$ and $n_0$ respectively denote the total count of attribute labels that equal 1 and 0.

We plot the estimated dictionary entry corresponding to the feature with the highest absolute correlation for selected attributes in Figure~\ref{fig:celeba_dict_corr}. Noting that features can be positive or negative for a given input data, it can be seen that the sparse priors lead to dictionary entries with features that visually resemble the semantic meaning of each attribute label.

%\begin{figure}[]
%\centering
    %\includegraphics[width=0.5\textwidth]{figures/celeba/recon.png}
    %\vspace{-0.3cm}
    %\caption{Sample reconstruction for models trained on CelebA with a DNN generator. From top to bottom, the rows represent the ground truth images, a standard VAE with a Gaussian prior, a sparse VAE with thresholded Gaussian+Gamma prior, and a sparse VAE with a Spike-and-Slab prior. Despite only using 10\% of latent features on average for each input data, the sparse VAEs have qualitatively similar reconstruction.}\label{fig:celeba_sample_recon}
%\end{figure}

\section{Discussion}
In this work, we have presented a new method for variational sparse coding by thresholding samples from an inference network. This simple approach avoids discrete random variables in its parameterization, and we show that it leads to superior performance and gradient estimation. Compared to a standard Gaussian prior, it provides more statistical efficiency in its use of latent features. One area of future work may consider alternative discrete random variable estimators that employ control variates \cite{tucker_rebar_2017}. Another interesting direction is the substitution of the KL divergence with distance metrics that perform better when the prior lies on a manifold (as is the case with sparse signals) \cite{patrini_sinkhorn_2020}.

\section{Acknowledgements}
The authors would like to thank Adam Charles, Francesco Tonolini, and Linxing Preston Jiang for their correspondence regarding their experience with variational sparse coding. This work was partially supported by NSF CAREER award CCF-1350954, ONR grant number N00014-15-1-2619 and AFRL grant number FA8750-19-C-0200.

% In the unusual situation where you want a paper to appear in the
% references without citing it in the main text, use \nocite
% \nocite{langley00}
\clearpage
\bibliography{main}
\bibliographystyle{icml2022}
\clearpage

%%%%%%%%%%%%%%%%%%%%%%%%%%%%%%%%%%%%%%%%%%%%%%%%%%%%%%%%%%%%%%%%%%%%%%%%%%%%%%%
%%%%%%%%%%%%%%%%%%%%%%%%%%%%%%%%%%%%%%%%%%%%%%%%%%%%%%%%%%%%%%%%%%%%%%%%%%%%%%%
% DELETE THIS PART. DO NOT PLACE CONTENT AFTER THE REFERENCES!
%%%%%%%%%%%%%%%%%%%%%%%%%%%%%%%%%%%%%%%%%%%%%%%%%%%%%%%%%%%%%%%%%%%%%%%%%%%%%%%
%%%%%%%%%%%%%%%%%%%%%%%%%%%%%%%%%%%%%%%%%%%%%%%%%%%%%%%%%%%%%%%%%%%%%%%%%%%%%%%
\appendix

\section{Thresholded Laplacian Derivation}
\label{sec:appendix_threshold_der}
To determine the distribution of our thresholded samples $p(z)$, we marginalize the conditional distribution with respect to $p(s)$. The shifted soft-threshold can be viewed as a shift of the pdf of $p(s)$ towards the mean value $\mu$, as depicted in Figure~\ref{fig:threshold_fig}.

We thus break up the marginalization over $p(s)$ into two disjoint regions: the region that gets collapsed into the origin $p(|s - \mu| \leq \lambda)$ and the regions that get shifted towards the mean $p(|s - \mu| \geq \lambda)$. We can then write the expectation:
\begin{align*}
p(z) = & \; \mathbb{E}_{p(s)} \left[ p(z \mid s)\right]\\
 = & \; p(z \mid |s - \mu| \leq \lambda)p(|s - \mu| \leq \lambda) \\
& + p(z \mid |s - \mu_0| \geq \lambda)p(|s - \mu| \geq \lambda) \\
= & \; (1 - \gamma) \delta(z) + \gamma p(z \mid | s - \mu| \geq \lambda)
\end{align*}
Where $(1 - \gamma) = 1 - \exp(-\lambda b^{-1})$ and $\gamma = \exp(-\lambda b^{-1})$, as shown in the main text.

To derive $p(z \mid |s - \mu| \geq \lambda)$, we write the pdf of $p(s)$ with the shift applied from soft-thresholding:
\begin{align*}
& \left \{
\begin{array}{ll}
      \frac{1}{2b}\exp\left(-\frac{|s - \mu + \lambda|}{b}\right) & s - \mu \leq 0 \\
      \frac{1}{2b}\exp\left(-\frac{|s - \mu  - \lambda|}{b}\right) & s - \mu > 0 \\
\end{array} \right.
\\
= & \frac{1}{2b}\exp\left(-\frac{|s - \mu + \mbox{sign}(s - \mu)\lambda|}{b}\right) \\
= & \frac{1}{2b}\exp\left(-\frac{|s - \mu|}{b}\right) \exp\left(-\lambda b^{-1}\right)
\end{align*}
Where the third line follows from the equality condition of the triangle inequality (both terms have the same sign) and from the fact that $\lambda$ is always positive.

This is precisely equal to $p(z \mid |s - \mu| \geq \lambda)$ multiplied by a normalizing constant $\frac{1}{Z}$ to become a valid distribution (accounting for the probability mass that was moved to zero). This normalizing constant is exactly one minus the mass moved to zero $Z = \gamma = \exp\left(-\lambda b^{-1}\right)$. This leads to our final conditional probability:
\begin{equation}
p(z \mid |s - \mu_0| \geq \lambda) = \frac{1}{2b}\exp\left(-\frac{|s - \mu|}{b}\right)
\end{equation}

%\section{Training with Shifted Soft-threshold Subgradient}

\section{Reparameterization Details for Thresholded Samples}
\label{sec:appendix_reparam}
To reparameterize thresholded samples, we first draw samples $s^{k}_i$ for each feature space dimension from a base distribution that is either Gaussian or Laplacian. For an input data $\mathbf{x}^k$, our encoder network outputs a shift and log-scale for each dimension of our latent feature for both distributions. The KL divergence loss is always computed on the base distribution, summing over each feature dimension.

For a Gaussian distribution, we apply the standard reparameterization technique from \cite{kingma_auto-encoding_2014} using $(\mu^k_i, 2 \log \sigma^k_i)$ output from a DNN:
\begin{equation*}
\epsilon^{j,k}_i \sim \mathcal{N}(0, 1)
\end{equation*}
\begin{equation*}
z^{j,k}_i = g_{\phi}(\mathbf{x}^k, \epsilon^{j,k}_i)_i = \sigma^k_i \epsilon^{j,k}_i + \mu^k_i
\end{equation*}
with a KL divergence depending on scale prior $\sigma^2_0$ \cite{kingma_auto-encoding_2014}:
\begin{equation*}
D_{KL}^{\mathcal{N}} = \biggl(\left(\mu^k_i\right)^2 + \left(\sigma^k_i\right)^2 \biggr)\biggl(2 \sigma_0\biggr)^{-1} + \log\frac{\sigma_0}{\sigma^k_i} - \frac{1}{2}
\end{equation*}

For a Laplacian distribution, we apply the reparameterization rule from \cite{connor_variational_2020} using $(\mu^k_i, \log b^k_i)$ output from a DNN:
\begin{equation*}
\epsilon^{j,k}_i \sim \mbox{Unif}\left(-\frac{1}{2}, \frac{1}{2}\right)
\end{equation*}
\begin{equation*}
z^{j,k}_i = g_{\phi}(\mathbf{x}^k, \epsilon^{j,k}_i)_i = -b \;\mbox{sign}(\epsilon^{j,k}_i) \log\left(1 - 2\epsilon^{j,k}_i \right)
\end{equation*}
Applying the KL divergence loss derived in \cite{meyer_alternative_2021} with a scale prior $b_0$:
\begin{equation*}
D_{KL}^{L} = \frac{|\mu^k_i|}{b_0} + \frac{b^k_i \exp\left(-|\mu^k_i|\left(b^k_i\right)^{-1}\right)}{b_0} + \log\frac{b_0}{b^k_i} - 1
\end{equation*}
To prevent numerical instability during training, we apply clamping to the log term in  $g_{\phi}(\mathbf{x}^k, \epsilon^{j,k}_i)_i$ to have a minimum value of $1\mathrm{E}
{-06}$. Additionally, we multiply a warm-up variable $\omega$ to the inferred scale variable $b^k_i$. At the beginning of training we set this variable $\omega^{(0)} = 0.1$ and increment it at each training iteration by $2\mathrm{E}{-04}$ until it reaches a maximum value of $1.0$.

When applying thresholding to either distribution, we either apply a fixed threshold set as a hyper-parameter $\lambda_i = \lambda_0$ or infer a threshold for each feature dimension. To perform reparameterization, we follow the pathwise derivative methods proposed in \cite{jankowiak_pathwise_2018}, parameterized by a neural network that outputs $(\log \alpha_i^k, \log \beta_i^k)$. For numerical stability, we clamp these values to be between $[1\mathrm{E}\mathrm{-06}, 1\mathrm{E}\mathrm{+06}]$. We use both the reparameterization and KL divergence implementations included in PyTorch 1.10 \cite{NEURIPS2019_9015}. In all cases, we set the prior $\alpha_0 = 3, \beta_0 = \frac{3}{\lambda_0}$. We find that on average this encourages the inferred threshold values to equal $\lambda_0$.

Our reparameterization procedure differs from the Spike-and-Slab from \cite{tonolini_variational_2020} in a few ways. First and foremost, we avoid the need of applying continuous approximations to Bernoulli random variables by applying shifted soft-thresholding. This also results in our model not needing to estimate a spike probability for each latent variable, but rather just parameters for the base distribution. Finally, rather than use the KL divergence for Spike-and-Slabs derived in \cite{tonolini_variational_2020}, we only penalize the base distribution with a KL divergence.

\section{Straight-Through Estimator Details}
\label{sec:appendix_subg}
The shifted soft-threshold is non-differentiable, with a subgradient of $0$ when a latent component $z_i = 0$. This is problematic since it serves as a block of gradient flowing from the generator to the inference network whenever a latent feature is not used (a common occurrence when sparsity is imposed on latent features). In these circumstances, although the gradient is blocked from the generator, a gradient is still computed for the KL divergence term. This empirically causes numerical instability (gradients go to NaN) or the inference network to primarily minimize the KL divergence term, leading to posterior collapse \cite{lucas_dont_2019}. Note that this same problem would occur with a reparameterization procedure that simply sets $z_i = 0$ without applying $\mathcal{T}_{\boldsymbol{\lambda}}$ in the cases where $\mathcal{T}_{\boldsymbol{\lambda}}$ would result in a zero-valued latent component.

Posterior collapse measures the extent to which the inference network is indistinguishable from the prior \cite{lucas_dont_2019}:
\begin{equation}
    \mathbb{P}_{\mathbf{x}\sim p(\mathbf{x})} \left[ D_{KL}\left( q_{\phi}(z_i, \mathbf{x}) \Vert p(z_i) \right) \leq \epsilon \right] \geq 1 - \delta.
\end{equation}
To measure another tangible problem of all latent features values going to zero, we also introduce the metric of feature collapse to measure whether inferred latent features $\widehat{\mathbf{z}}$ are zero for most images $\mathbf{x}$:
\begin{equation}
    \mathbb{P}_{\mathbf{x}\sim p(\mathbf{x})} \left[ |\widehat{z_i}| \leq \epsilon \right] \geq 1 - \delta.
\end{equation}

We measure these two quantities in Table~\ref{tab:posterior_collapse}  using subgradient and straight-through estimators. It can be observed that using a single sample, the subgradient leads to numerical instability in all three runs. Increasing the number of samples avoids this instability, at the cost of complete posterior and feature collapse in all three runs. Meanwhile, the straight-through estimator avoids this issue.

\begin{table*}[t]
\centering
\caption{\label{tab:posterior_collapse} Posterior collapse and feature collapse for different sparsity estimators. Thresholding methods use a straight-through (ST) estimator or a subgradient (SG) estimator. NaN denotes a run that failed due to gradients equal to NaN.}
\begin{tabular}{@{}ccccc@{}}
\toprule
Method/Prior Distribution & \multicolumn{2}{c}{\% Posterior Collapse} & \multicolumn{2}{c}{\% Feature Collapse}  \\
 & \multicolumn{2}{c}{$(\epsilon = 1\mathrm{E}\mathrm{-02}, \delta = 5\mathrm{E}\mathrm{-02})$}   & \multicolumn{2}{c}{$(\epsilon = 1\mathrm{E}\mathrm{-02}, \delta = 5\mathrm{E}\mathrm{-02})$}  \\
 & J=1 & J=20 & J=1 & J=20 \\
\midrule
%Spike-and-slab              & 0.52\% & 0.13\% & 0.00\% & 0.00\%\\
Thresholded Laplacian (SG)       & NaN & 100.00\% & NaN & 100.00\%\\
Thresholded Laplacian (ST)       & 4.17\% & 4.56\% & 0.00\% & 0.00\%\\
\bottomrule
\end{tabular}
\end{table*}

\section{Linear Generator Training Details}
\label{sec:linear_train_detail}
For each method, we train using $80,000$ image patches and validate using $16,000$ image patches of dimension $16 \times 16$ with a dictionary of $256$ entries. We train for $300$ epochs using a batch size of $100$. Our initial learning rate for the dictionary is $5\mathrm{E}\mathrm{-01}$ and we apply an exponential decay by a factor of $0.99$ each epoch. Our inference network is trained with an initial learning rate of $1\mathrm{E}\mathrm{-02}$, using an SGD+Nesterov optimizer with a CycleScheduler. For FISTA, we set $\kappa = 1\mathrm{E}{-03}$. For the variational methods, we set $\kappa = 1\mathrm{E}{-04}$, and the scale parameter of our prior $p(\mathbf{z})$ equal to $0.1$ when applicable. Unless specified, we set the KL weight as $\beta = 1\mathrm{E}\mathrm{-02}$ for each method. We run each method for three trials, using the random seeds $\{0, 1337, 747 \}$.

For FISTA, we multiply $\lambda$ by a warmup parameter $\omega$ that starts at $\omega^{(0)} = 0.1$ and is incremented by $1\mathrm{E}\mathrm{-04}$ each training iteration, capping at a maximum value of $\omega^{(t)} = 1.0$. We find that this prevents all dictionary entries from going to zero when training with a Frobenius norm regularizer.

For the Spike-and-Slab, we use the same warmup strategy proposed in \cite{tonolini_variational_2020}. We set $\omega^{(0)} = 0.0$, incrementing by  $2\mathrm{E}\mathrm{-04}$ each iteration after iteration $1500$. This parameter is also capped out at $\omega^{(t)} = 1.0$. Additionally, we apply a warmup to the temperature parameter $\tau$ for the Gumbel-Softmax used in the Spike-and-Slab. This starts at $\tau^{(0)} = 1.0$ and is decreased by a multiplicative factor of $0.9995$ until it reaches a minimum value of $\tau^{(t)} = 5\mathrm{E}\mathrm{-01}$. We set the prior for the spike probability as $0.1$.

For the Thresholded Gaussian and Laplacian methods, we use the analytic results to set $\boldsymbol{\lambda}_0$ respectively as $0.52$ and $0.25$. We find that these values lead to $10\%$ of the latent features non-zero for each input data on average. When using a Gamma prior on the threshold parameter, we reduce the corresponding KL weight by a multiplicative factor of $1\mathrm{E}\mathrm{-01}$ (resulting in $\beta_1 = 1\mathrm{E}\mathrm{-02}$ and $\beta_2 = 1\mathrm{E}\mathrm{-03}$). We set $\boldsymbol{\alpha}_0 = 3$ for each method that uses a Gamma prior.

Each inference network uses an MLP backbone with the same architecture, with a separate projection head for each distribution parameter. The architecture for the backbone is outlined in Table~\ref{tab:lin_gen_backbone}. For each distribution parameter, an additional $256 \times 256$ linear layer is added on top of the output of this network.

\begin{table}[h]
\centering
\caption{Inference Network Backbone for Linear Generator}
\vspace{0.5em}
\label{tab:lin_gen_backbone}
\begin{tabular}{||l||}
 \hline
 Input $\in \mathbb{R}^{16 \times 16}$ \\
 Linear: $256 \times 512$   \\
 ReLU \\
 Linear: $512 \times 1024$   \\
 ReLU \\
 Linear: $1024 \times 512$   \\
 ReLU \\
 Linear: $512 \times 256$   \\
 ReLU \\
 Output $\in \mathbb{R}^{256}$ \\
\hline
\end{tabular}
\end{table}

\section{DNN Generator Training Details}
\label{sec:dnn_train_detail}
For CelebA, we use $150,000$ training samples and $19,000$ validation samples. All samples are center cropped to $140 \times 140$ pixels and then resized to $64 \times 64$. We also apply a random horizontal flip to training samples only. We train for $300$ epochs using a batch size of $512$ across two Nvidia RTX 3080s. We use an initial learning rate of $3\mathrm{E}\mathrm{-04}$ using the Adam optimizer with $\beta = (0.5, 0.999)$, weight decay equal to $1\mathrm{E}\mathrm{-05}$, and a sample budget of $J=10$. Additionally, we use the automatic mixed precision (AMP) and DistributedDataParallel implementations included in PyTorch 1.10~\cite{NEURIPS2019_9015}. We set $\beta = 1\mathrm{E}\mathrm{-03}$ for the KL divergence term in all methods. In the case where we have a Gamma distribution on the threshold parameter, we multiply the KL term by a factor of $1\mathrm{E}\mathrm{-01}$ (leading to $\beta_1 = 1\mathrm{E}\mathrm{-03}$ and $\beta_2 = 1\mathrm{E}\mathrm{-04}$). The network architecture for the encoder and the decoder are included in Table~\ref{tab:DNN_gen}. All other methodology is kept consistent with the linear generator. We run each method for three trials, using the random seeds $\{0, 1337, 747 \}$.

For FMNIST, we train with all $60,000$ training samples and evaluate on all $10,000$ validation samples. Hyper-parameters are kept the same as CelebA experiments, except for a KL divergence weight $\beta = 1\mathrm{E}\mathrm{-02}$ and sampling budget $J=20$. A smaller network architecture is used for FMNIST, depicted in Table~\ref{tab:FMNIST_DNN_gen}.

\begin{table*}[!htb]
\centering
\caption{Network Architecture for CelebA Experiments}
\label{tab:DNN_gen}
\begin{tabular}{||l l||}
 \hline
 Encoder Network & Decoder Network  \\
 \hline
 Input $\in \mathbb{R}^{64 \times 64 \times 3}$ & Input $\in \mathbb{R}^d$  \\
 conv: chan: 64, kern: 4, stride: 2, pad: 1  & Linear: $d \times 1024$ Units  \\
 BatchNorm: feat: 64 & ReLU\\
 ReLU &  convTranpose: chan: 128, kern: 4, stride: 2, pad: 1\\
 conv: chan: 64, kern: 4, stride: 2, pad: 1  & BatchNorm: feat: 128 \\
 BatchNorm: feat: 64 & ReLU\\
 ReLU &  convTranpose: chann: 128, kern: 4, stride: 2, pad: 1 \\
 conv: chan: 128, kern: 4, stride: 2, pad: 1 &  BatchNorm: feat: 128\\
  BatchNorm: feat: 128 & ReLU \\
 ReLU & convTranpose: chan: 64, kernel: 4, stride: 2, pad: 1\\
 conv: chan: 256, kern: 4, stride: 2, pad: 1 &  BatchNorm: feat: 64\\
  BatchNorm: feat: 256 & ReLU \\
 ReLU & convTranpose: chan: 64, kernel: 4, stride: 2, pad: 1\\
 conv: chan: 256, kern: 4, stride: 2, pad: 0 &  BatchNorm: feat: 64\\
  BatchNorm: feat: 256 & ReLU \\
 ReLU & convTranpose: chan: 3, kernel: 4, stride: 2, pad: 1\\
  & Sigmoid \\
 Output $\in \mathbb{R}^{256}$ & Output $\in \mathbb{R}^{64 \times 64 \times 3}$\\

 \hline
\end{tabular}
\end{table*}

\begin{table*}[!htb]
\centering
\caption{Network Architecture for FMNIST Experiments}
\label{tab:FMNIST_DNN_gen}
\begin{tabular}{||l l||}
 \hline
 Encoder Network & Decoder Network  \\
 \hline
 Input $\in \mathbb{R}^{28 \times 28 \times 1}$ & Input $\in \mathbb{R}^d$  \\
 conv: chan: 64, kern: 4, stride: 2, pad: 1  & Linear: $d \times 128$ Units  \\
 BatchNorm: feat: 64 & ReLU\\
 ReLU &  convTranpose: chan: 128, kern: 4, stride: 2, pad: 0\\
 conv: chan: 64, kern: 4, stride: 2, pad: 1  & BatchNorm: feat: 128 \\
 BatchNorm: feat: 64 & ReLU\\
 ReLU &  convTranpose: chann: 64, kern: 4, stride: 1, pad: 0 \\
 conv: chan: 128, kern: 4, stride: 1, pad: 0 &  BatchNorm: feat: 64\\
  BatchNorm: feat: 128 & ReLU \\
 ReLU & convTranpose: chan: 64, kernel: 4, stride: 2, pad: 1\\
 conv: chan: 64, kern: 4, stride: 2, pad: 1 &  BatchNorm: feat: 64\\
  BatchNorm: feat: 128 & ReLU \\
 ReLU & convTranpose: chan: 1, kernel: 4, stride: 2, pad: 1\\
  & Sigmoid \\
 Output $\in \mathbb{R}^{128}$ & Output $\in \mathbb{R}^{28 \times 28 \times 1}$\\

 \hline
\end{tabular}
\end{table*}
\section{Additional Results for Linear Generator}
\label{sec:extra_linear_results}

\begin{table*}[t!]
\centering
\caption{\label{tab:quant_linear_std} A copy of Table~\ref{linear_gen_comp} from the main text with $J=20$ samples with standard deviations included}
\begin{tabular}{@{}cccc@{}}
\toprule
Method/Prior Distribution & Validation Loss   & Multi-Information & IWAE Loss                 \\
\midrule
FISTA (baseline)            & 1.01E+02 $\pm$ 7.00E-01    & 8.75E+01 $\pm$ 5.43E-01 & --   \\
Gaussian                    & 1.35E+03 $\pm$ 2.08E+00 & 7.36E+02 $\pm$ 3.06E-01 & 2.16E-01 $\pm$ 2.64E-04  \\
Laplacian                   & 5.79E+02 $\pm$ 1.51E+00 & 5.34E+02 $\pm$ 5.29E-01 & 2.22E-01 $\pm$ 7.97E-04 \\
Spike-and-slab              & 2.39E+02 $\pm$ 4.27E+00 & 1.96E+02 $\pm$ 2.57E+00 & 2.12E-01 $\pm$ 2.64E-02 \\
Thresholded Gaussian        & 2.30E+02 $\pm$ 1.87E+00 & 1.76E+02 $\pm$ 2.08E-01 & 1.33E+00 $\pm$ 3.80E-03 \\
Thresholded Gaussian+Gamma  & 2.70E+02 $\pm$ 1.87E+00 & 2.29E+02 $\pm$ 3.79E-01 & 1.13E+00 $\pm$ 2.47E-03 \\
Thresholded Laplacian       & 1.94E+02 $\pm$ 1.66E+00 & 1.91E+02 $\pm$ 1.53E-01 & 1.13E+00 $\pm$ 1.29E-02 \\
Thresholded Laplacian+Gamma & 2.11E+02 $\pm$ 1.47E+00 & 2.33E+02 $\pm$ 5.77E-01 & 1.12E+00 $\pm$ 8.09E-03 \\    \bottomrule
\end{tabular}
\end{table*}

\begin{figure*}[h]
\centering
    \subfigure[Inference Network SNR]{\hspace{2.5em}\includegraphics[width=0.8\textwidth]{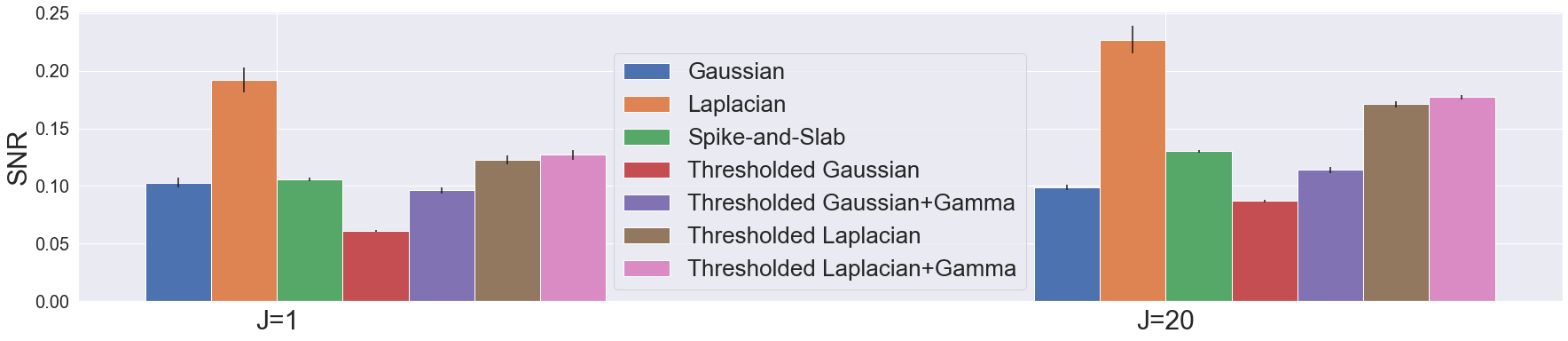}}
    \newline
    \subfigure[Generator SNR]{\includegraphics[width=0.8\textwidth]{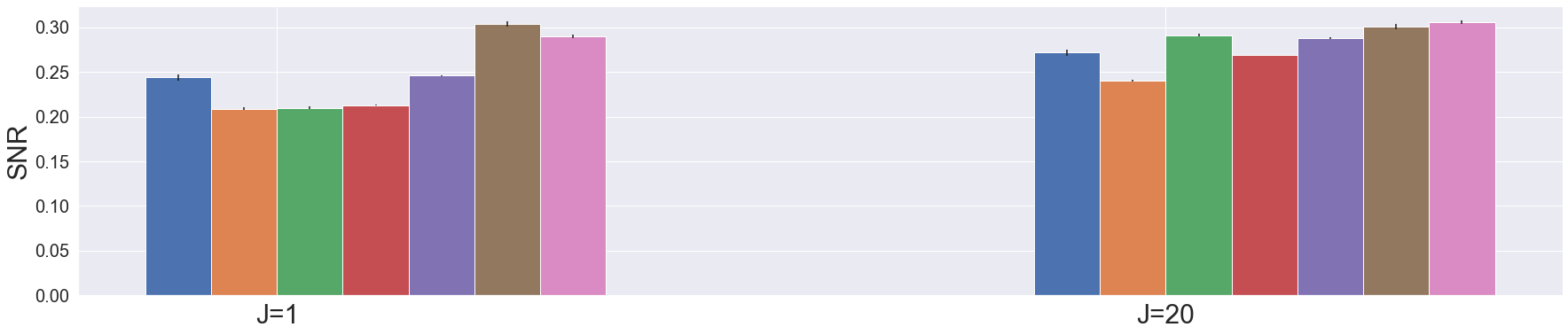}}

    \caption{Comparison of SNR for (a) inference network and (b) linear generator for all methods using $J=1$ and $J=20$ with max ELBO sampling.}
\end{figure*}

\subsection{Full Dictionary for Different Priors}
\begin{figure*}[h]
\centering
    \subfigure[]{\includegraphics[width=0.45\textwidth]{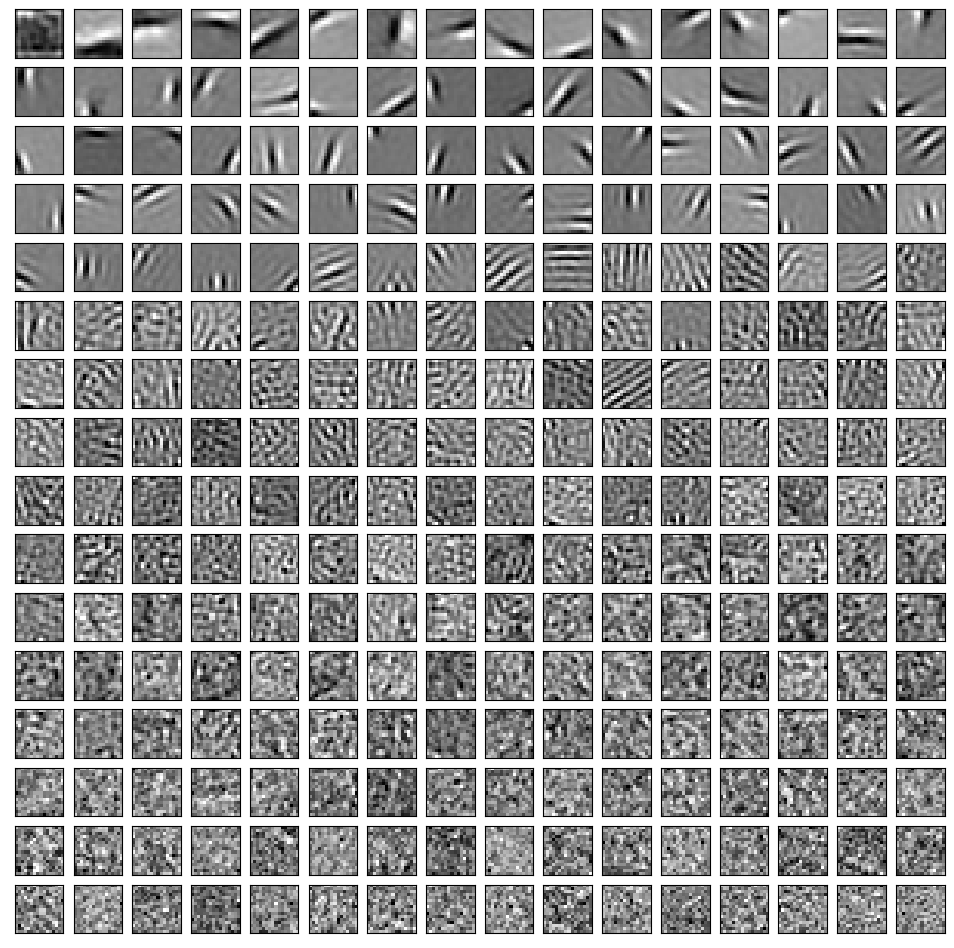}}
    ~
    \caption{Full dictionary sorted by magnitude in descending order for FISTA with $\lambda=20$ and a Frobenius norm penalty of $1\mathrm{E}\mathrm{-03}$.}
\end{figure*}
\begin{figure*}[h]
\centering
    \subfigure[$J=1$]{\includegraphics[width=0.45\textwidth]{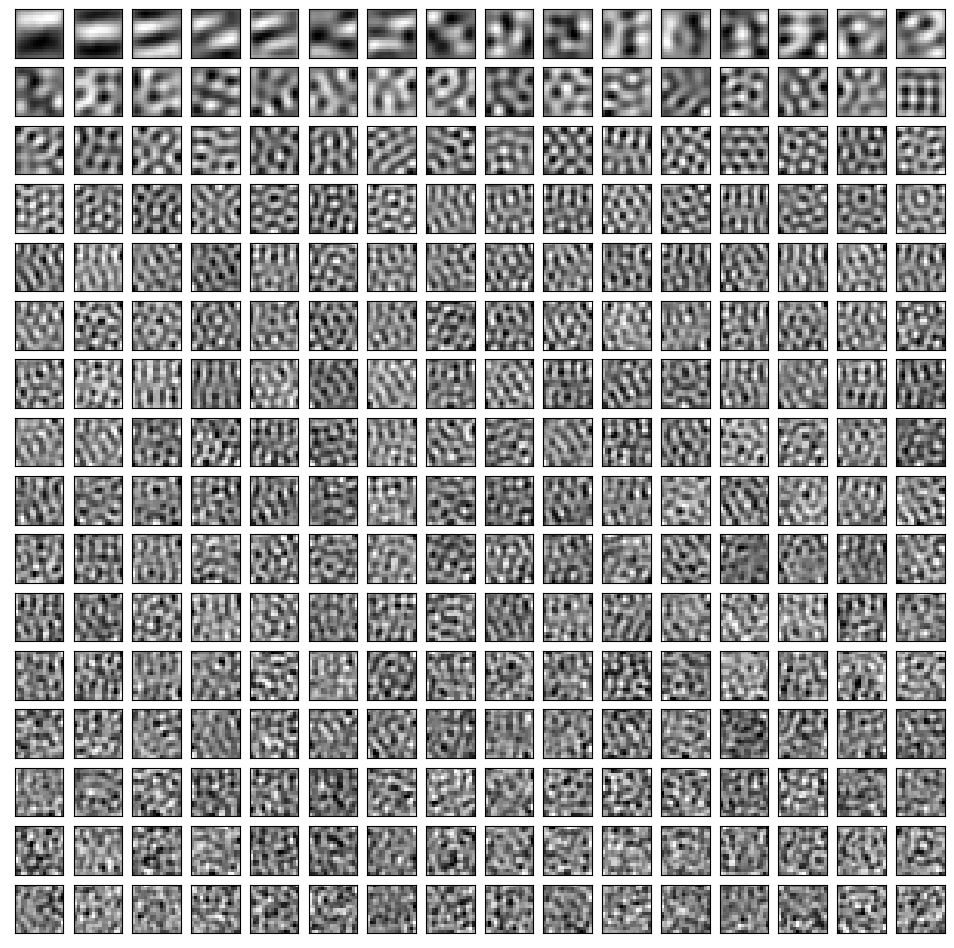}}
    ~
    \subfigure[$J=20$]{\includegraphics[width=0.45\textwidth]{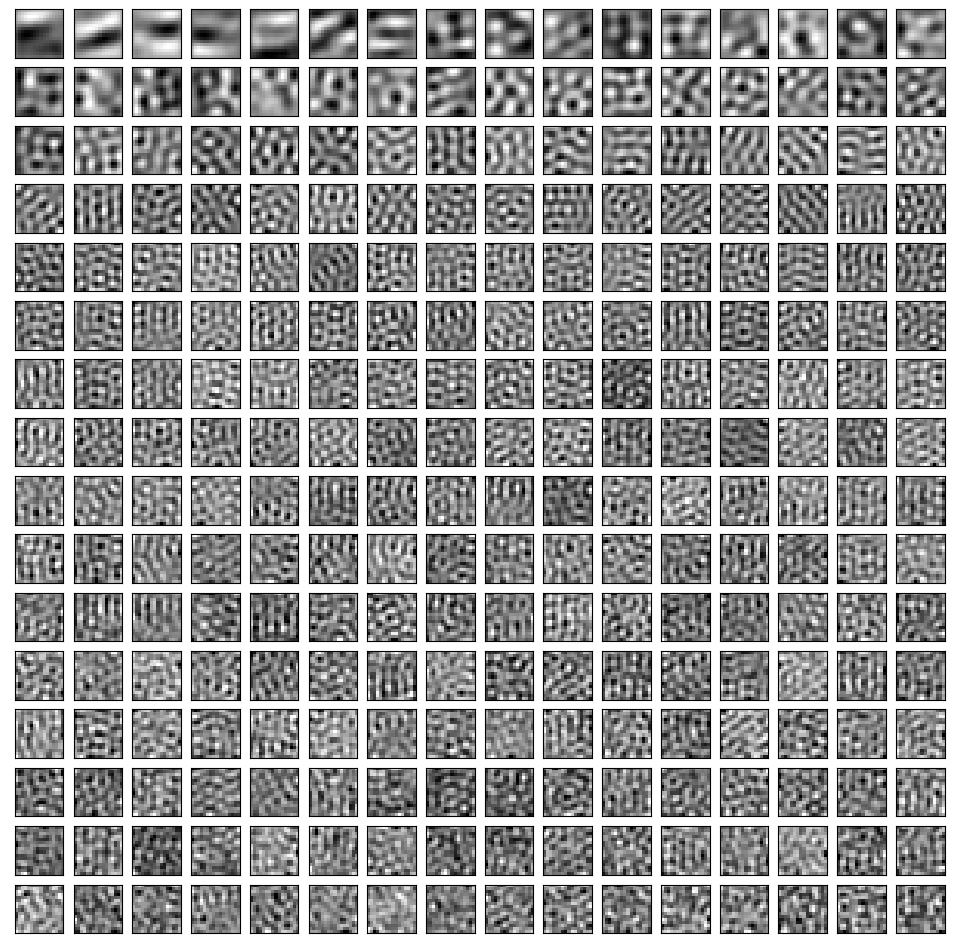}}
    ~
    \caption{Full dictionary sorted by magnitude in descending order for \textbf{Gaussian} prior with $J=1$ and $J=20$ under max ELBO sampling.}
\end{figure*}
\begin{figure*}[h]
\centering
    \subfigure[$J=1$]{\includegraphics[width=0.45\textwidth]{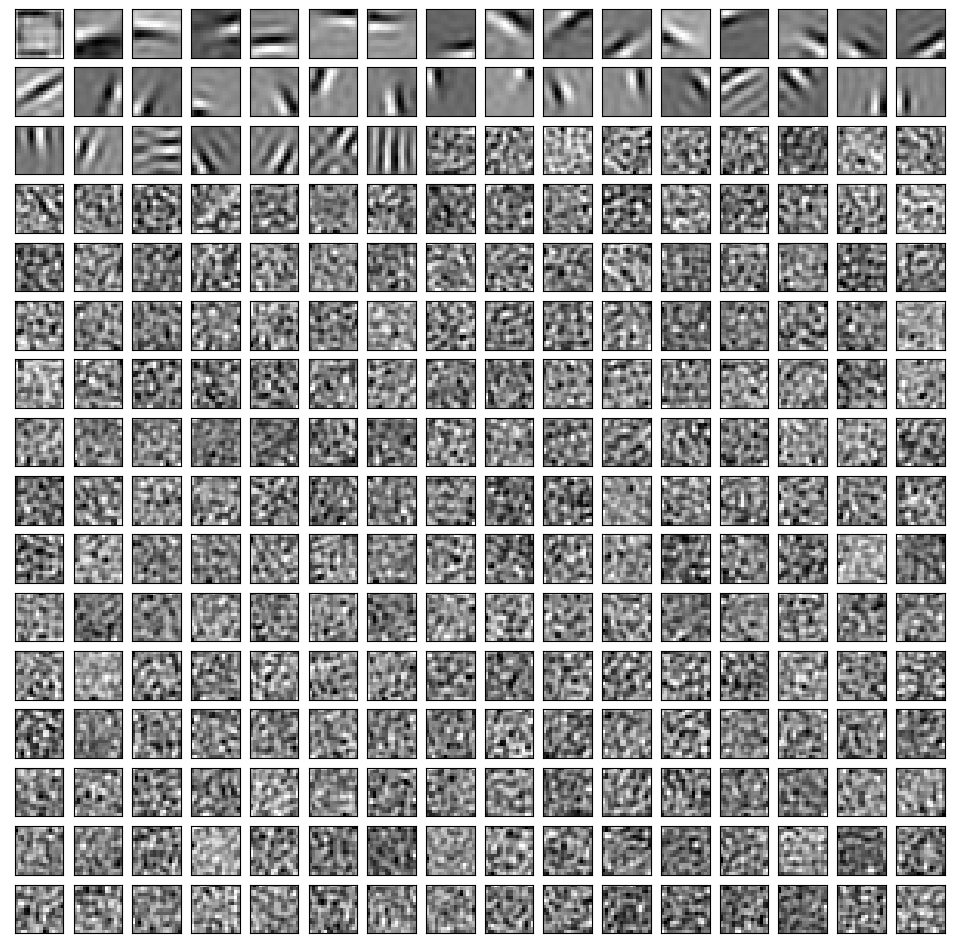}}
    ~
    \subfigure[$J=20$]{\includegraphics[width=0.45\textwidth]{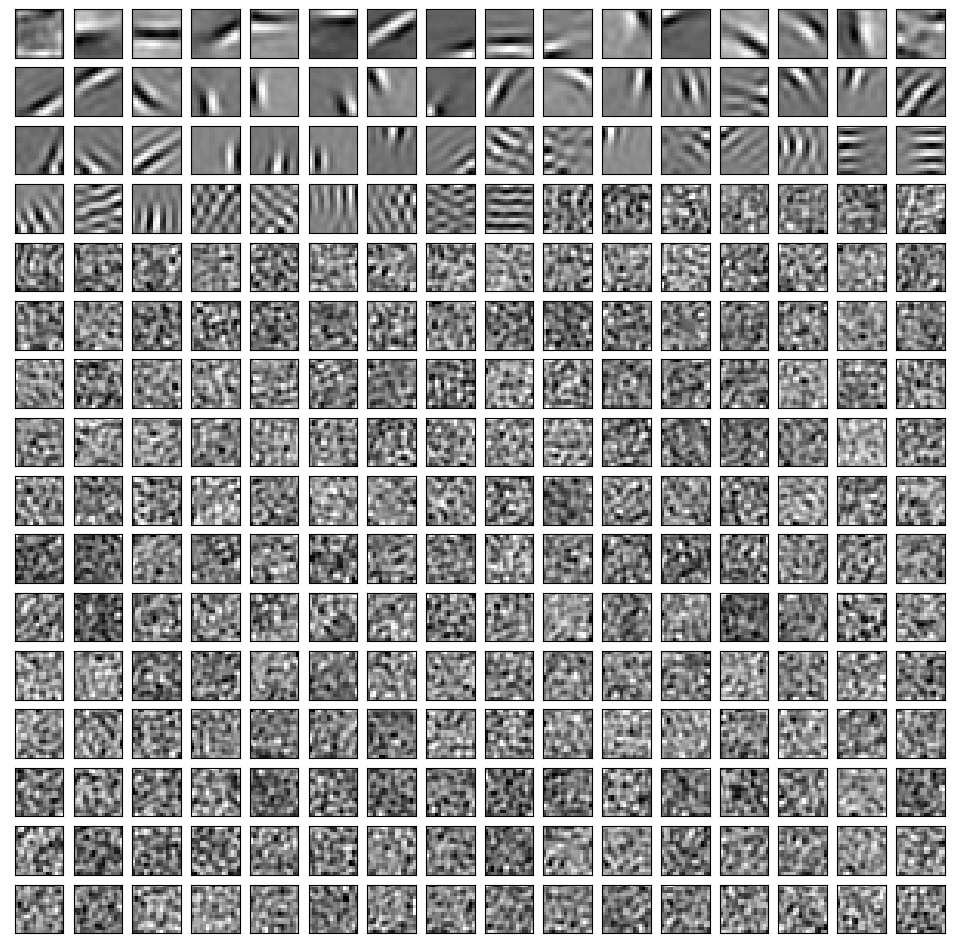}}
    ~
    \caption{Full dictionary sorted by magnitude in descending order for \textbf{Laplacian} prior with $J=1$ and $J=20$ under max ELBO sampling.}
\end{figure*}
\begin{figure*}[h]
\centering
    \subfigure[$J=1$]{\includegraphics[width=0.45\textwidth]{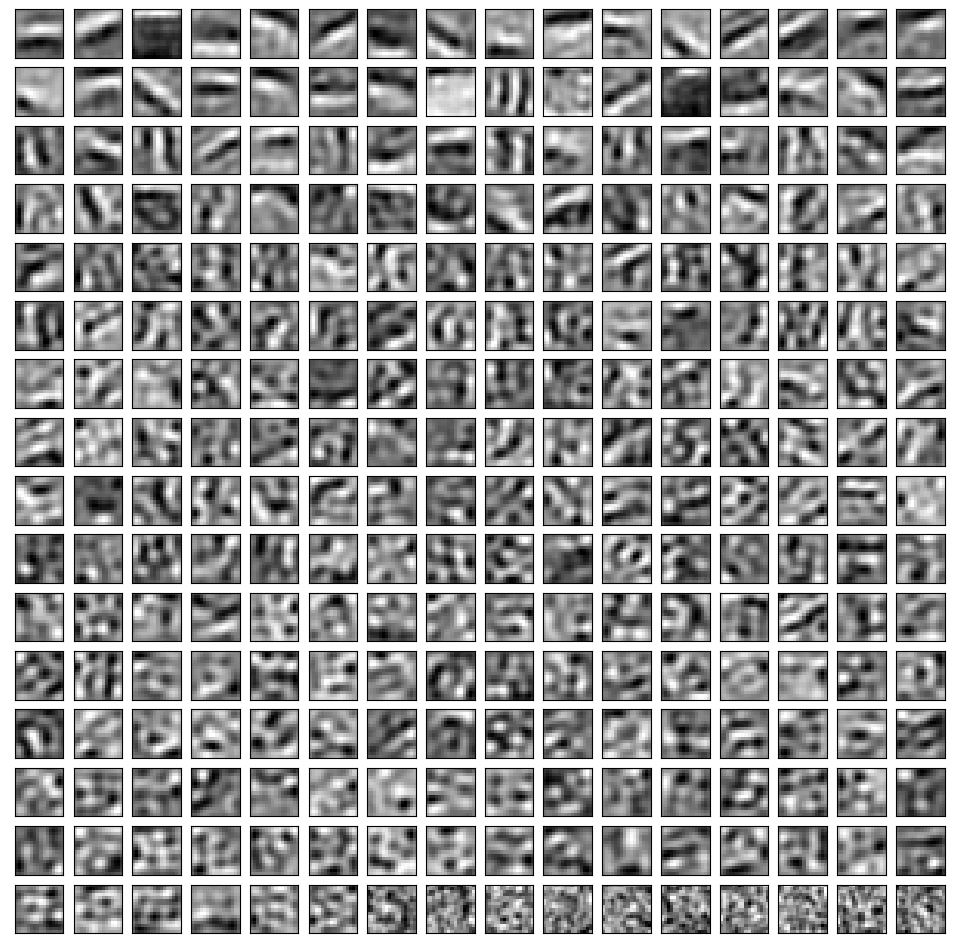}}
    ~
    \subfigure[$J=20$]{\includegraphics[width=0.45\textwidth]{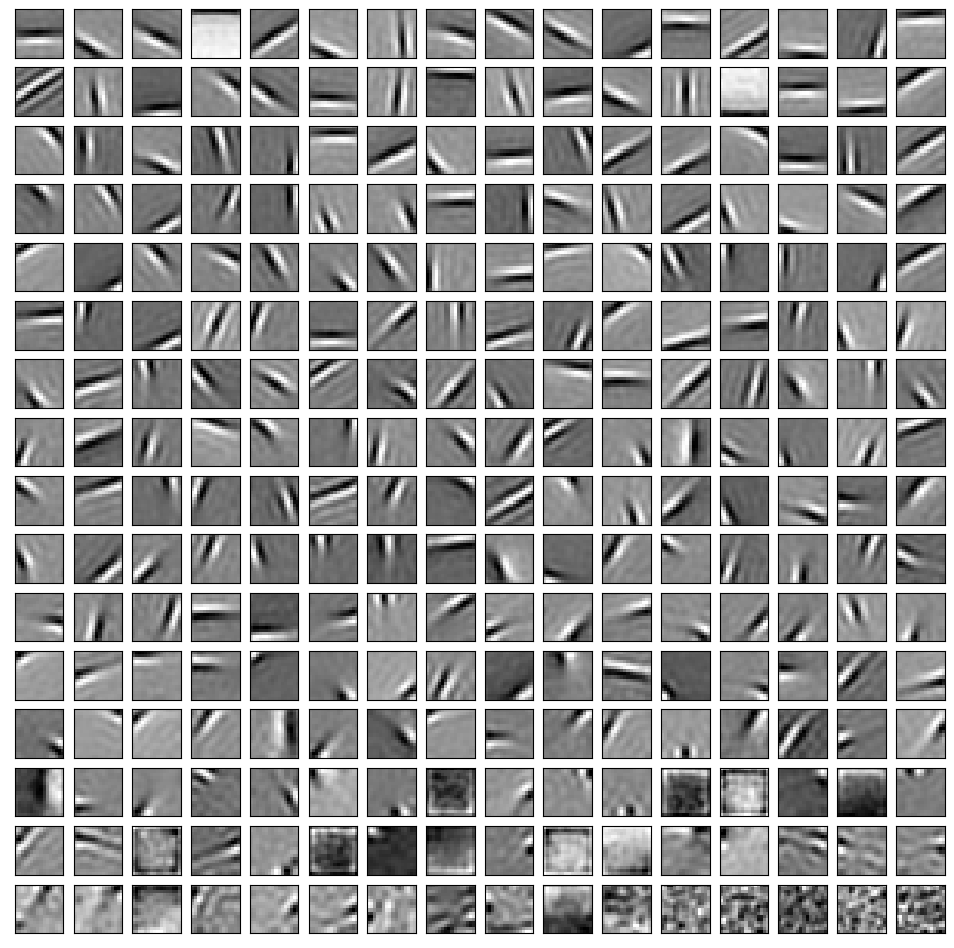}}
    ~
    \caption{Full dictionary sorted by magnitude in descending order for \textbf{Spike-and-Slab} prior with $J=1$ and $J=20$ under max ELBO sampling.}
\end{figure*}
\begin{figure*}[h]
\centering
    \subfigure[$J=1$]{\includegraphics[width=0.45\textwidth]{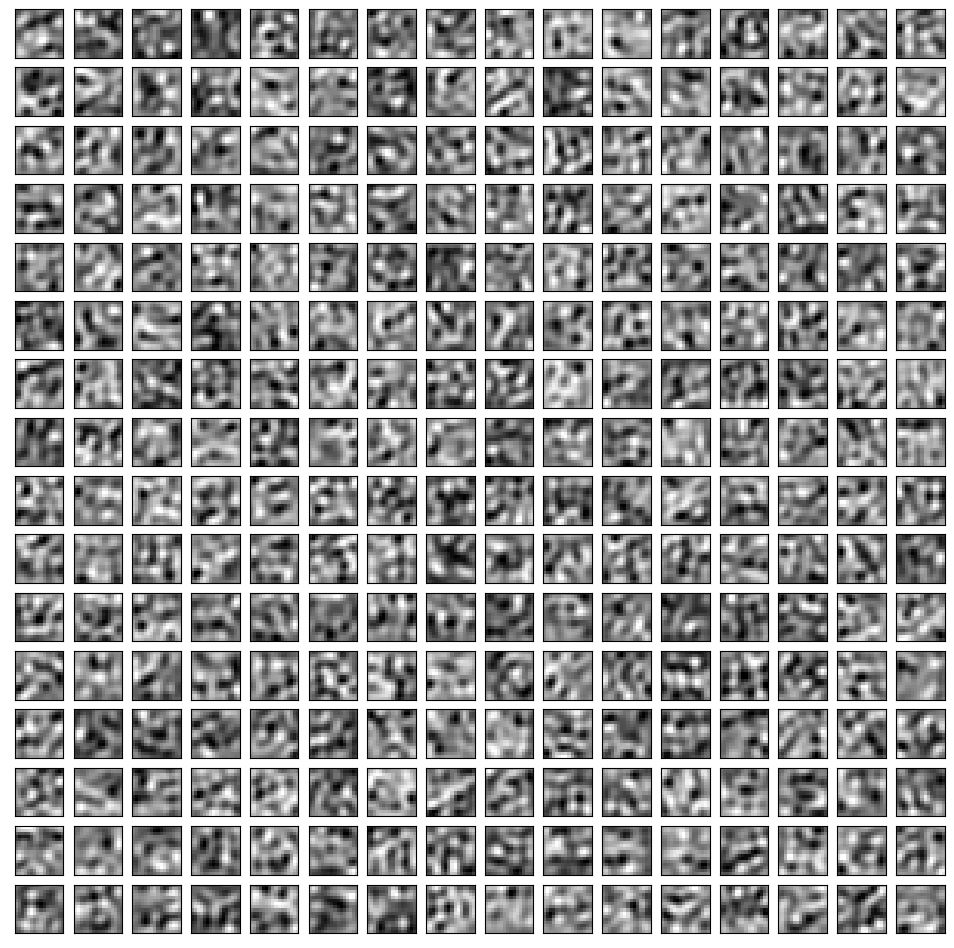}}
    ~
    \subfigure[$J=20$]{\includegraphics[width=0.45\textwidth]{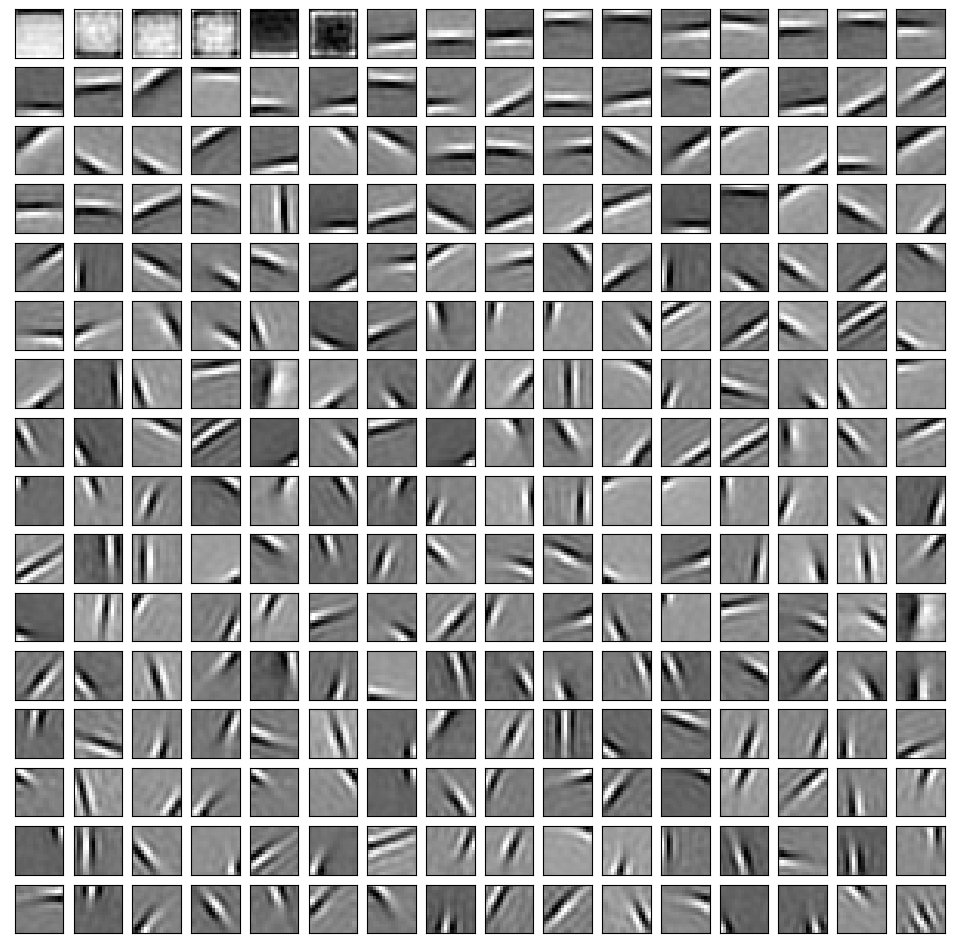}}
    ~
    \caption{Full dictionary sorted by magnitude in descending order for \textbf{Thresholded Gaussian} prior with $J=1$ and $J=20$ under max ELBO sampling.}
\end{figure*}
\begin{figure*}[h]
\centering
    \subfigure[$J=1$]{\includegraphics[width=0.45\textwidth]{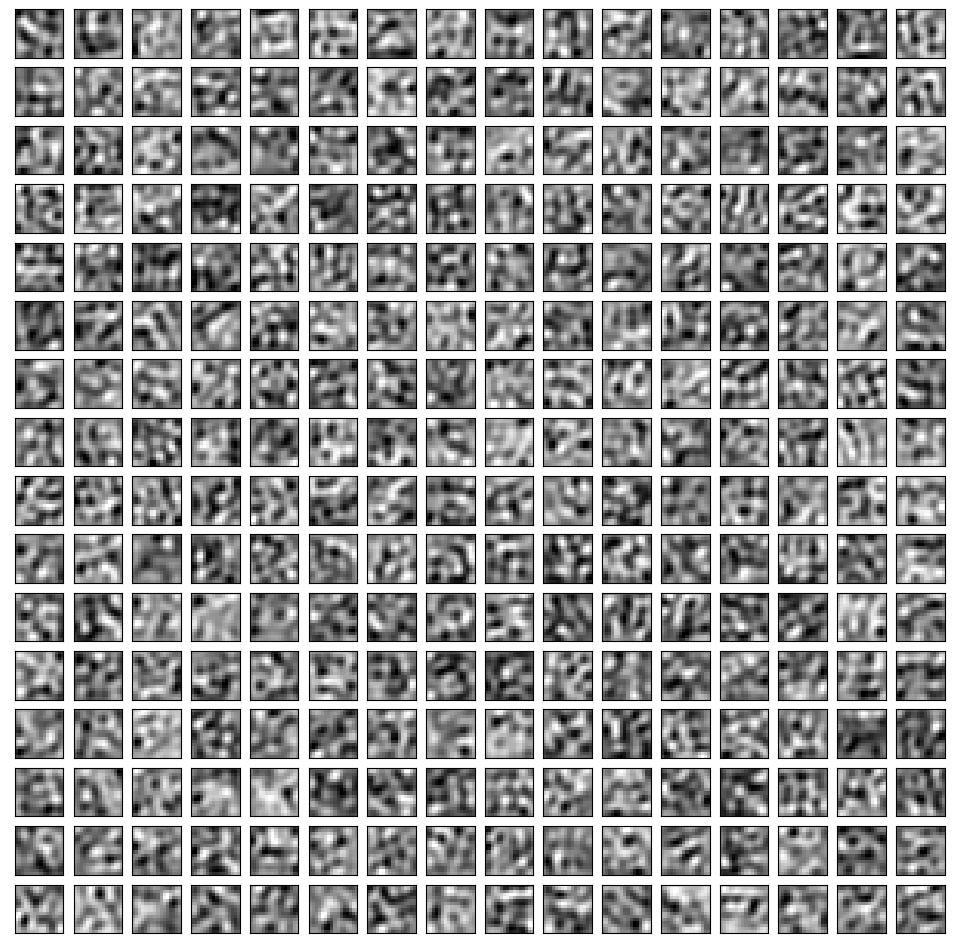}}
    ~
    \subfigure[$J=20$]{\includegraphics[width=0.45\textwidth]{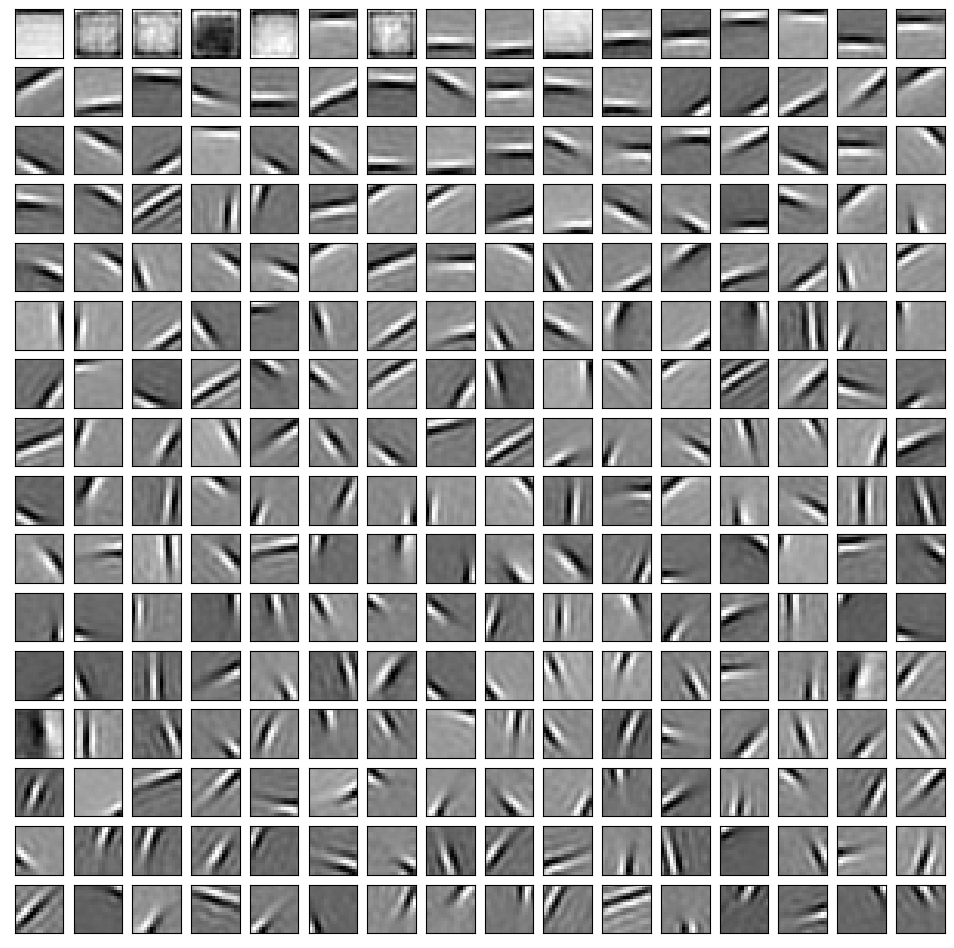}}
    ~
    \caption{Full dictionary sorted by magnitude in descending order for \textbf{Thresholded Gaussian + Gamma threshold prior} prior with $J=1$ and $J=20$ under max ELBO sampling.}
\end{figure*}
\begin{figure*}[h]
\centering
    \subfigure[$J=1$]{\includegraphics[width=0.45\textwidth]{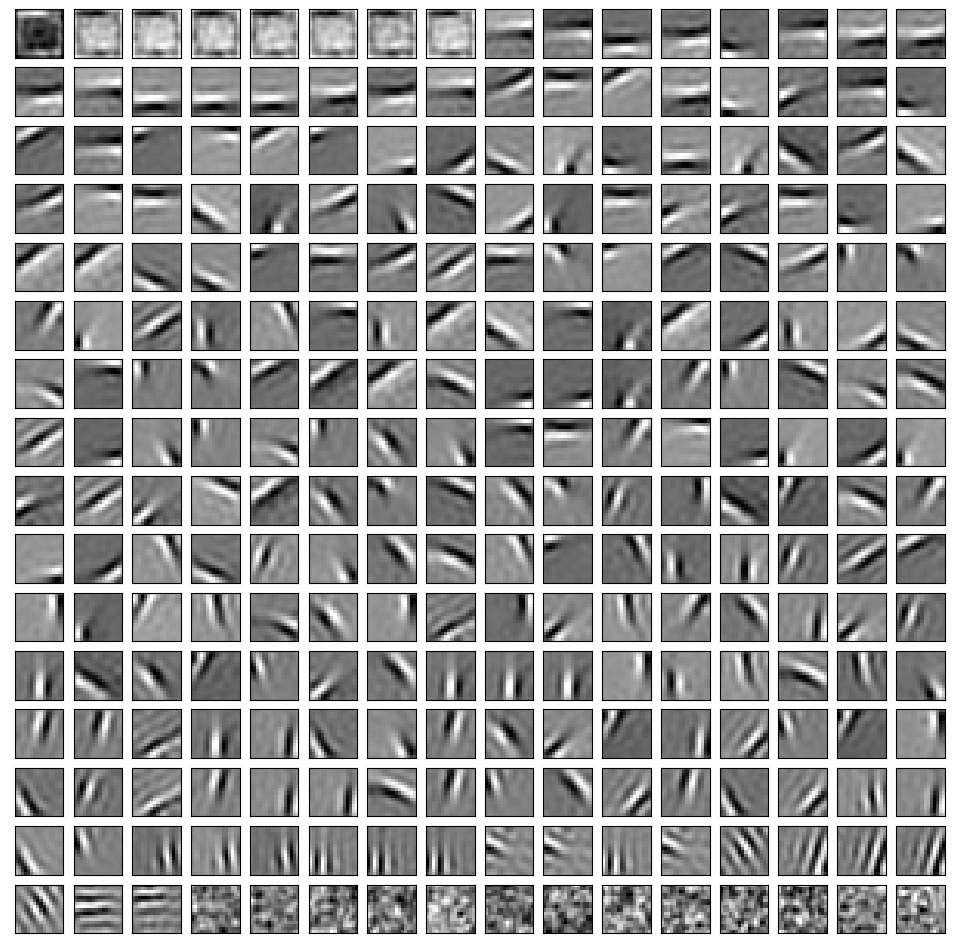}}
    ~
    \subfigure[$J=20$]{\includegraphics[width=0.45\textwidth]{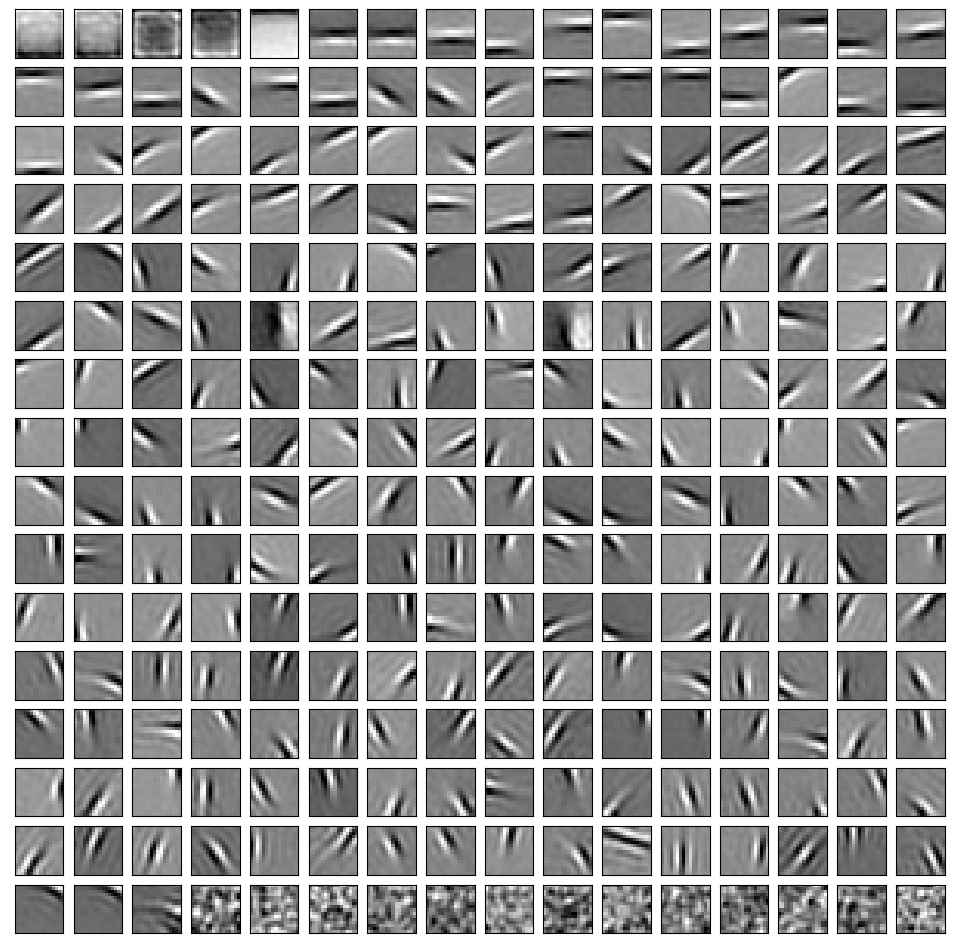}}
    ~
    \caption{Full dictionary sorted by magnitude in descending order for \textbf{Thresholded Laplacian} prior with $J=1$ and $J=20$ under max ELBO sampling.}
\end{figure*}
\begin{figure*}[h]
\centering
    \subfigure[$J=1$]{\includegraphics[width=0.45\textwidth]{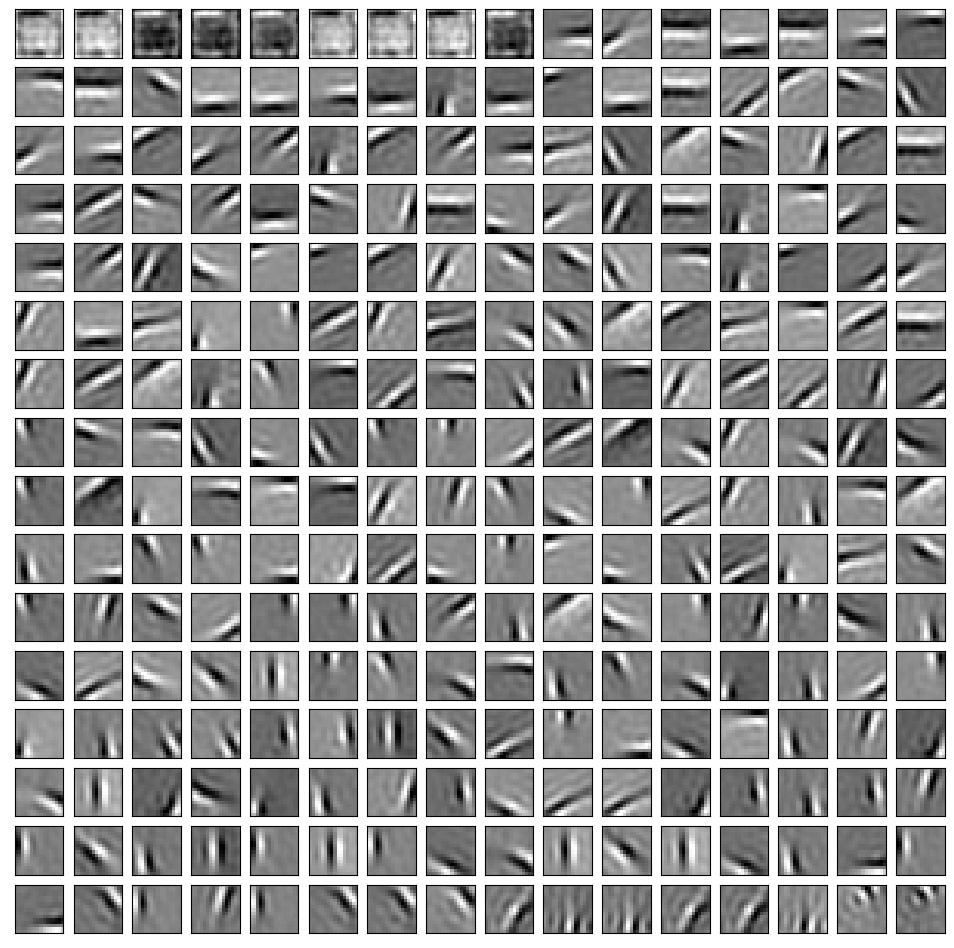}}
    ~
    \subfigure[$J=20$]{\includegraphics[width=0.45\textwidth]{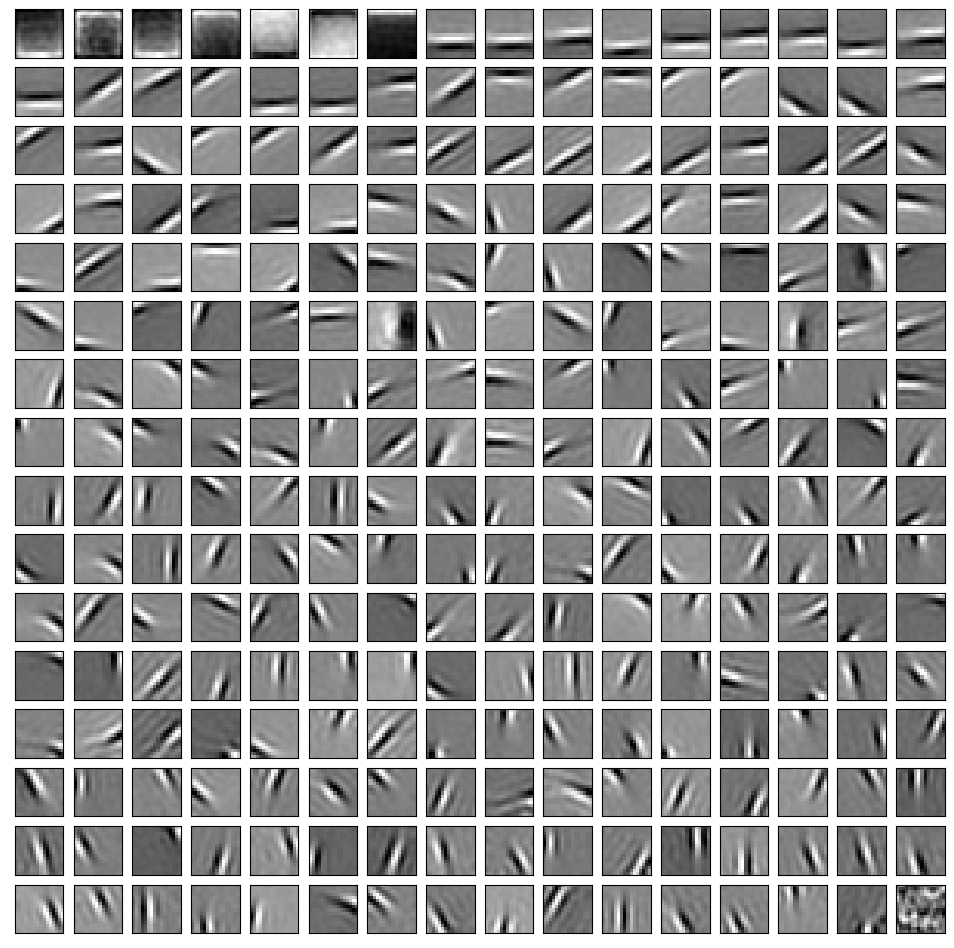}}
    ~
    \caption{Full dictionary sorted by magnitude in descending order for \textbf{Thresholded Laplacian + Gamma threshold prior} prior with $J=1$ and $J=20$ under max ELBO sampling.}
\end{figure*}
%\clearpage

\subsection{Full Dictionary for Different Sampling Methods}
\label{sec:full_dict_sampling}
\begin{figure*}[b!]
\label{fig:full_sample_dict}
\centering
    \subfigure[Avg]{\includegraphics[width=0.3\textwidth]{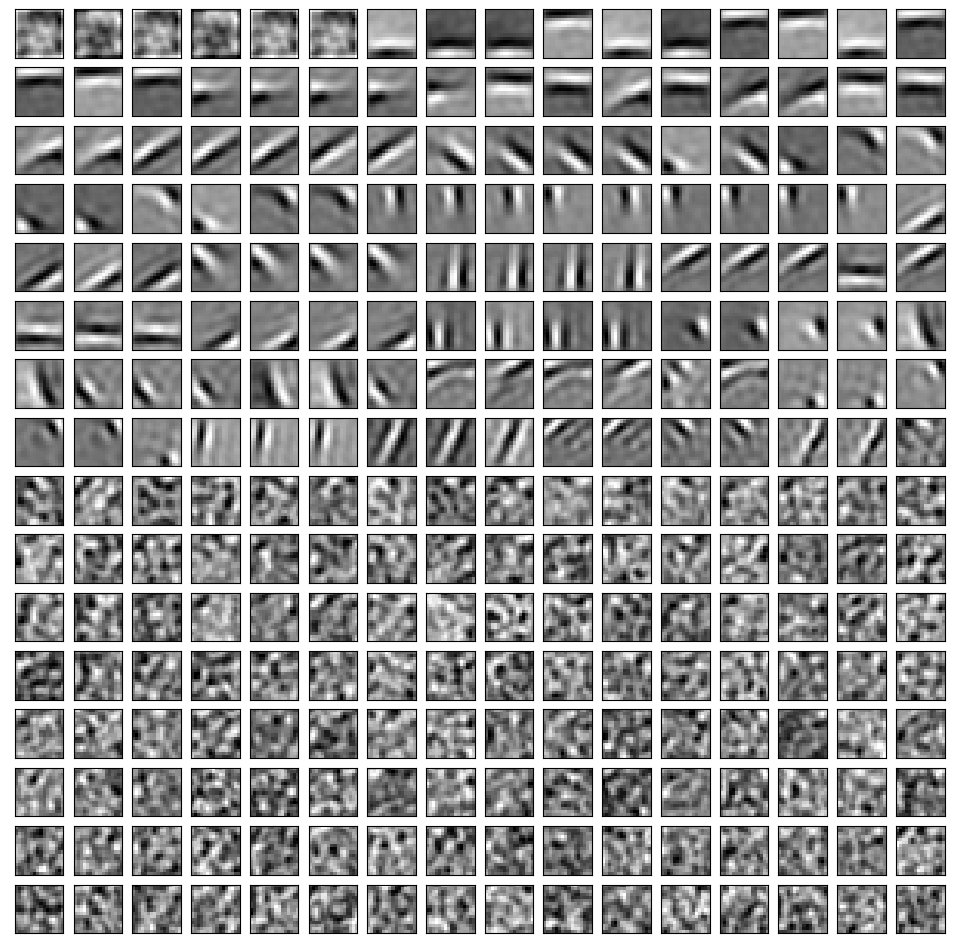}}
    ~
    \subfigure[IWAE]{\includegraphics[width=0.3\textwidth]{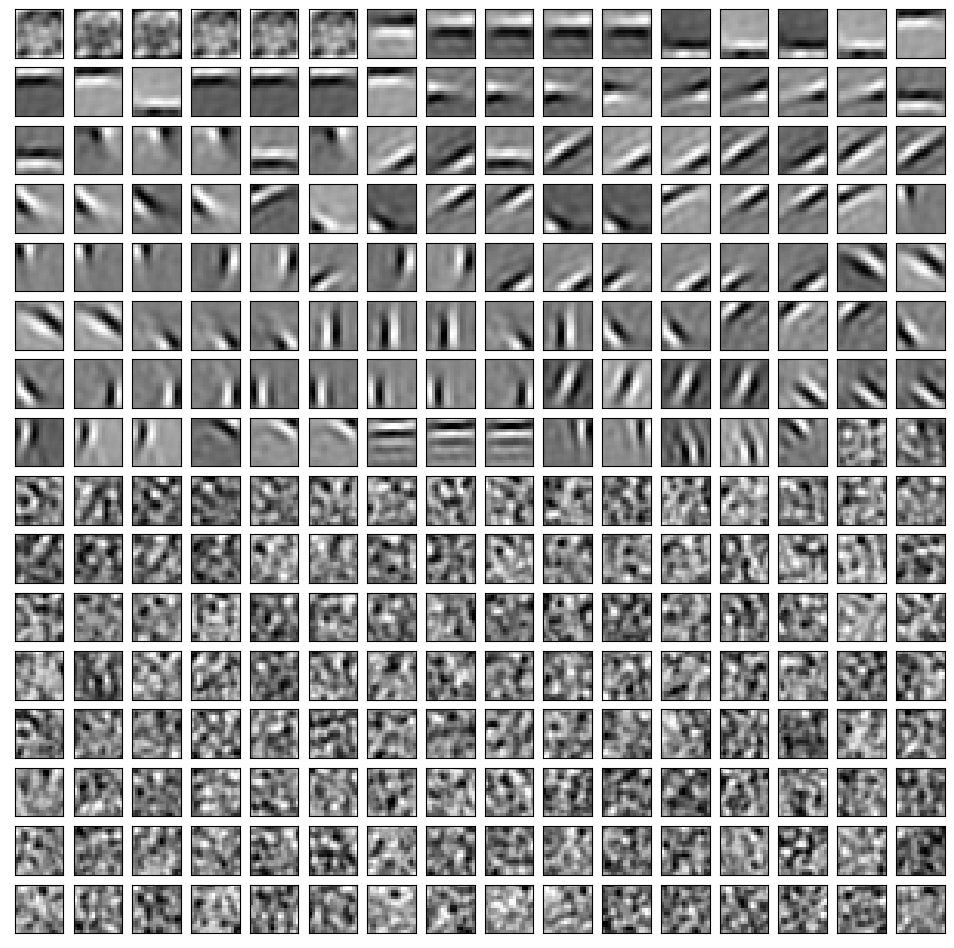}}
    ~
    \subfigure[Max]{\includegraphics[width=0.3\textwidth]{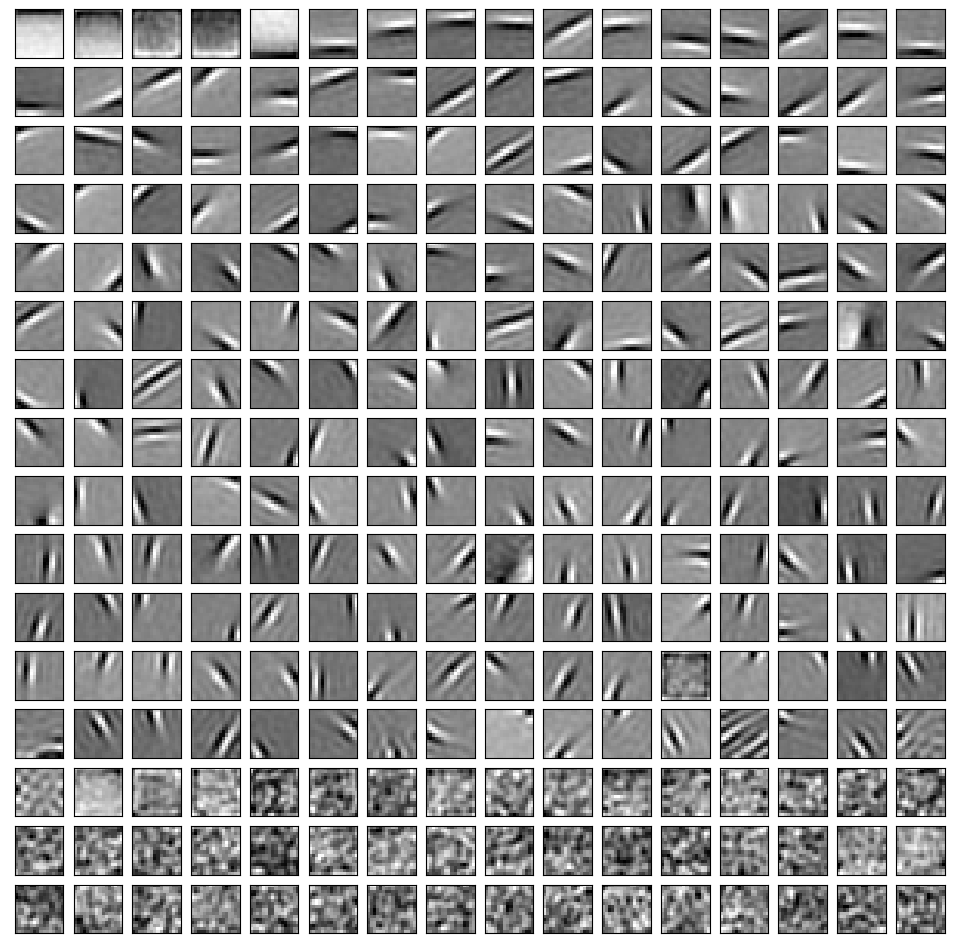}}
    ~
    \caption{Full dictionary sorted by magnitude in descending order for different sampling methods using $J=100$ samples using the Thresholded Laplacian prior. Average sampling and  IWAE sampling learn repetetive features in their dictionary since they do not encourage feature reuse during training. On the other hand, max ELBO sampling learns a larger dictionary of diverse features.}
\end{figure*}
%\clearpage

\section{Additional Results for DNN Generator}
\label{sec:dnn_extra_results}

\begin{figure*}[htb!]
\centering
    \subfigure[64 Dimensions]{\includegraphics[width=0.46\textwidth]{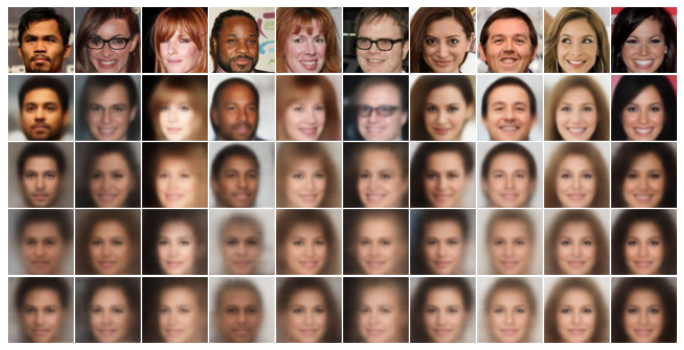}}
    ~
    \subfigure[4096 Dimensions]{\includegraphics[width=0.46\textwidth]{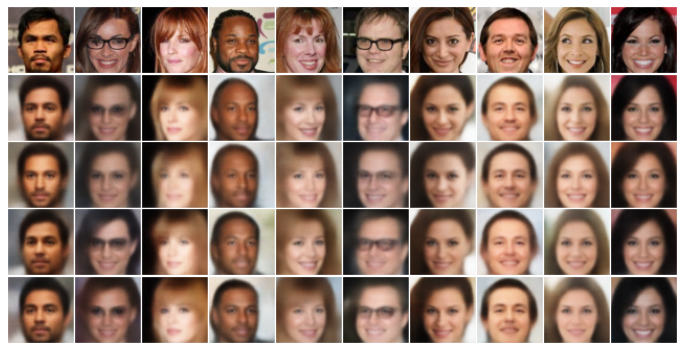}}
    ~
    \caption{Image reconstruction on CelebA using DNN generator for (a) 64 dimensions and (b) 4096 dimensions. From top to bottom, the rows denote the ground truth images, a Gaussian prior, a Spike-and-Slab prior, a Thresholded Gaussian prior, and a Thresholded Gaussian+Gamma prior.}
\end{figure*}

\subsection{Recovered Dictionary at Different Dimensionality}

\begin{figure*}[h]
\centering
    \subfigure[Gaussian]{\includegraphics[width=0.46\textwidth]{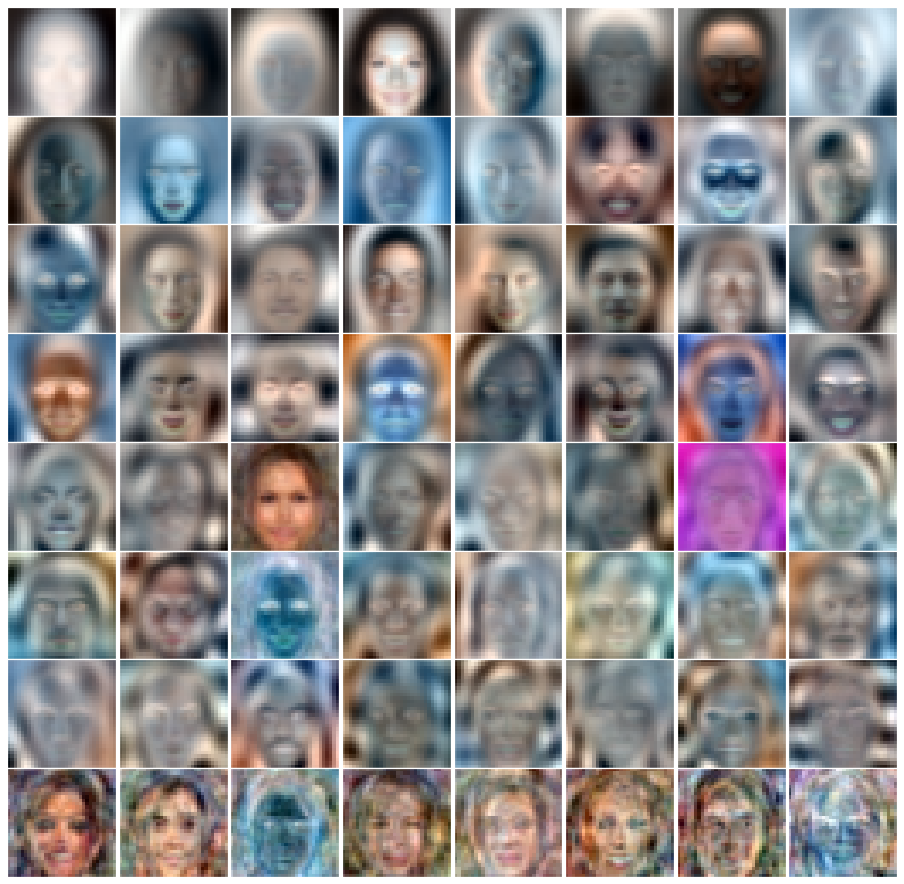}}
    ~
    \subfigure[Thresholded Gaussian+Gamma]{\includegraphics[width=0.46\textwidth]{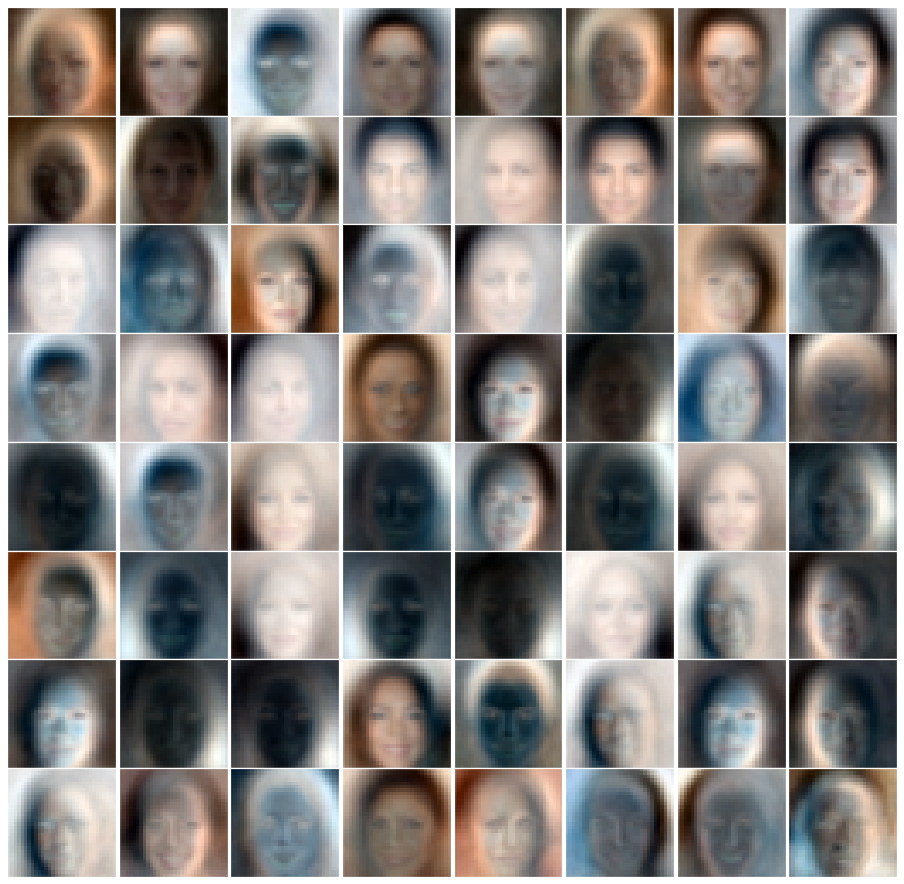}}
    ~
    \caption{\textbf{64 dimensional latent space}. Sampled entries from an estimated dictionary using an inference network trained with a DNN generator on CelebA. Entries are shown in descending order of Frobenius norm magnitude, with each row taking linearly spaced entries from the full dictionary. The sparse prior is set so to encourage 10\% of latent features to be non-zero.}
\end{figure*}

\begin{figure*}[h]
\centering
    \subfigure[Gaussian]{\includegraphics[width=0.46\textwidth]{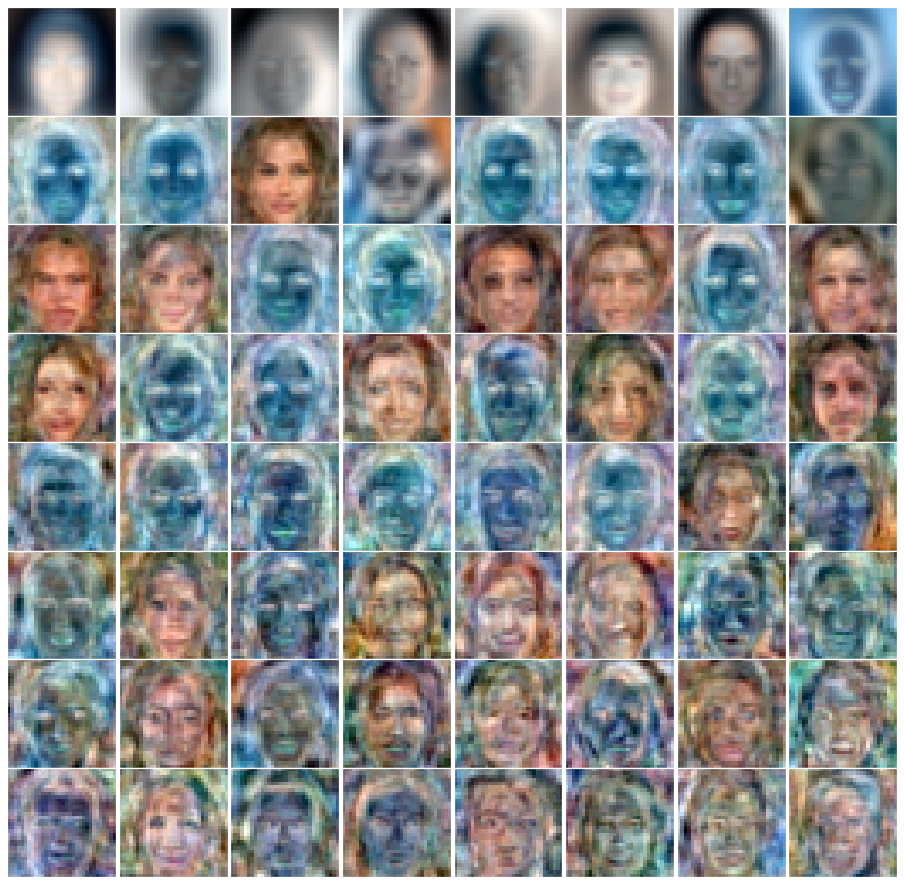}}
    ~
    \subfigure[Thresholded Gaussian+Gamma]{\includegraphics[width=0.46\textwidth]{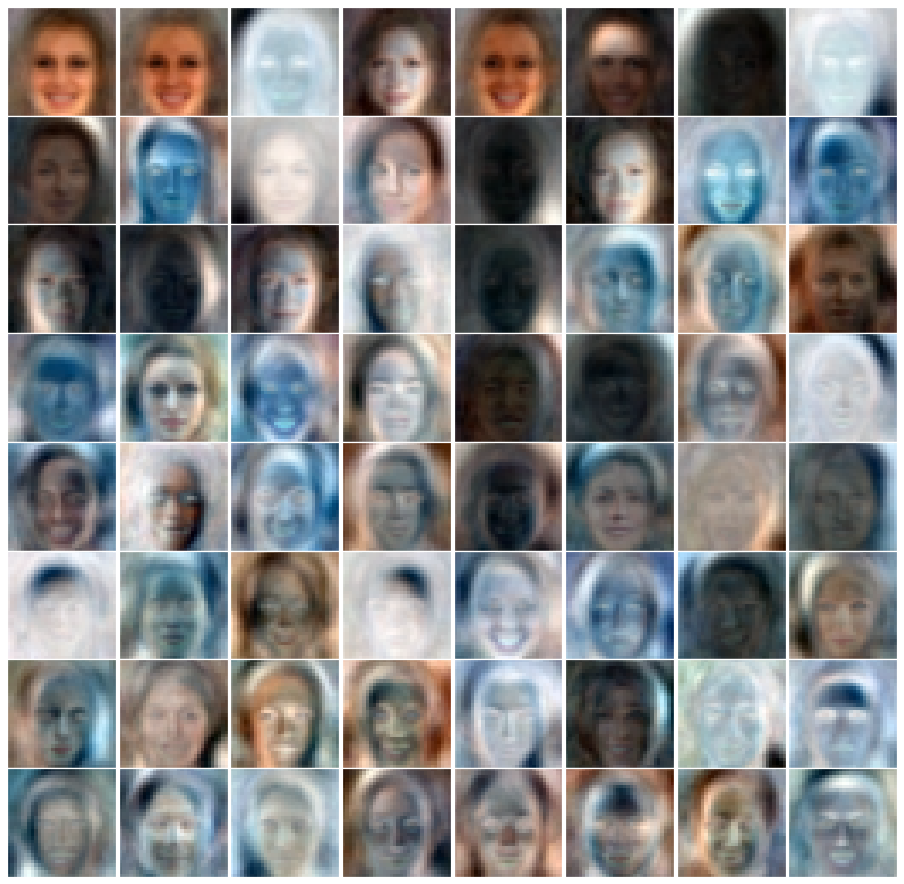}}
    ~
    \caption{\textbf{512 dimensional latent space}. Sampled entries from an estimated dictionary using an inference network trained with a DNN generator on CelebA. Entries are shown in descending order of Frobenius norm magnitude, with each row taking linearly spaced entries from the full dictionary. The sparse prior is set so to encourage 10\% of latent features to be non-zero.}
\end{figure*}

\begin{figure*}[h]
\centering
    \subfigure[Gaussian]{\includegraphics[width=0.46\textwidth]{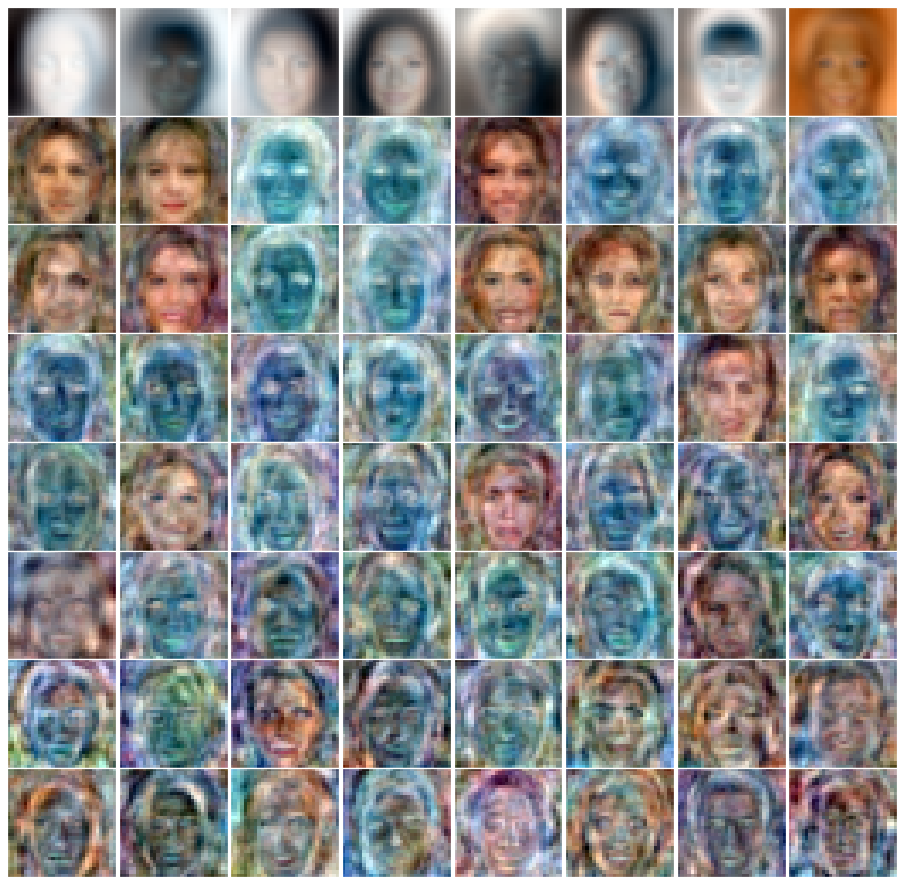}}
    ~
    \subfigure[Thresholded Gaussian+Gamma]{\includegraphics[width=0.46\textwidth]{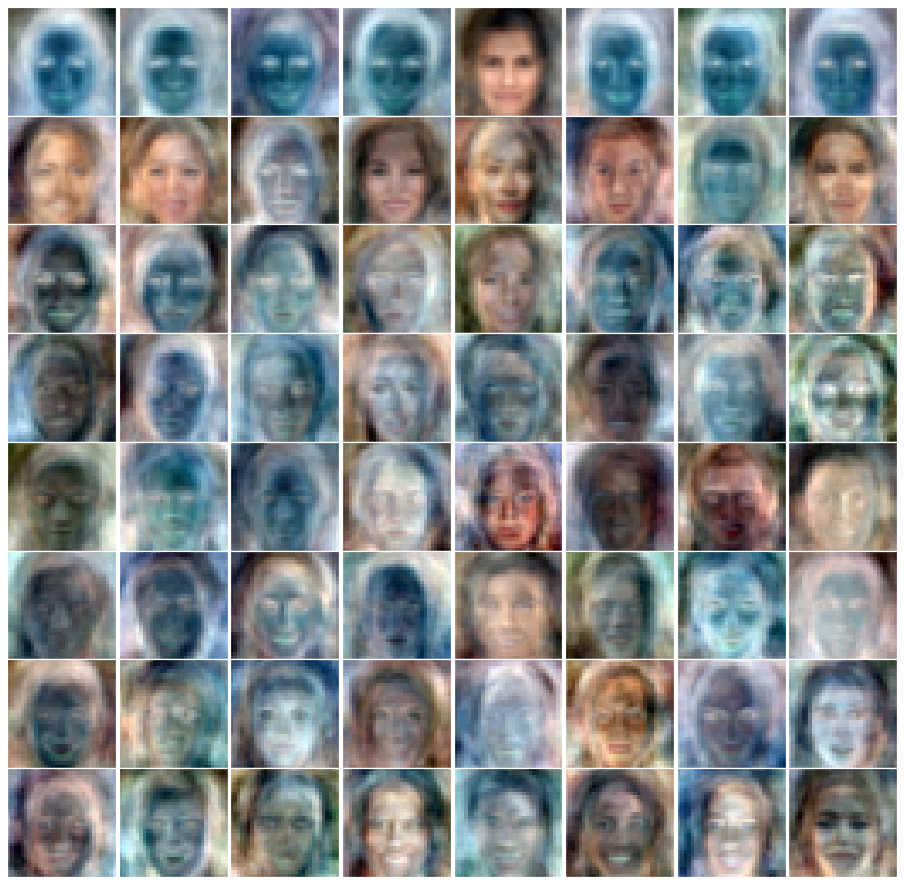}}
    ~
    \caption{\textbf{2048 dimensional latent space}. Sampled entries from an estimated dictionary using an inference network trained with a DNN generator on CelebA. Entries are shown in descending order of Frobenius norm magnitude, with each row taking linearly spaced entries from the full dictionary. The sparse prior is set so to encourage 10\% of latent features to be non-zero.}
\end{figure*}

\begin{figure*}[h]
\centering
    \subfigure[Gaussian]{\includegraphics[width=0.46\textwidth]{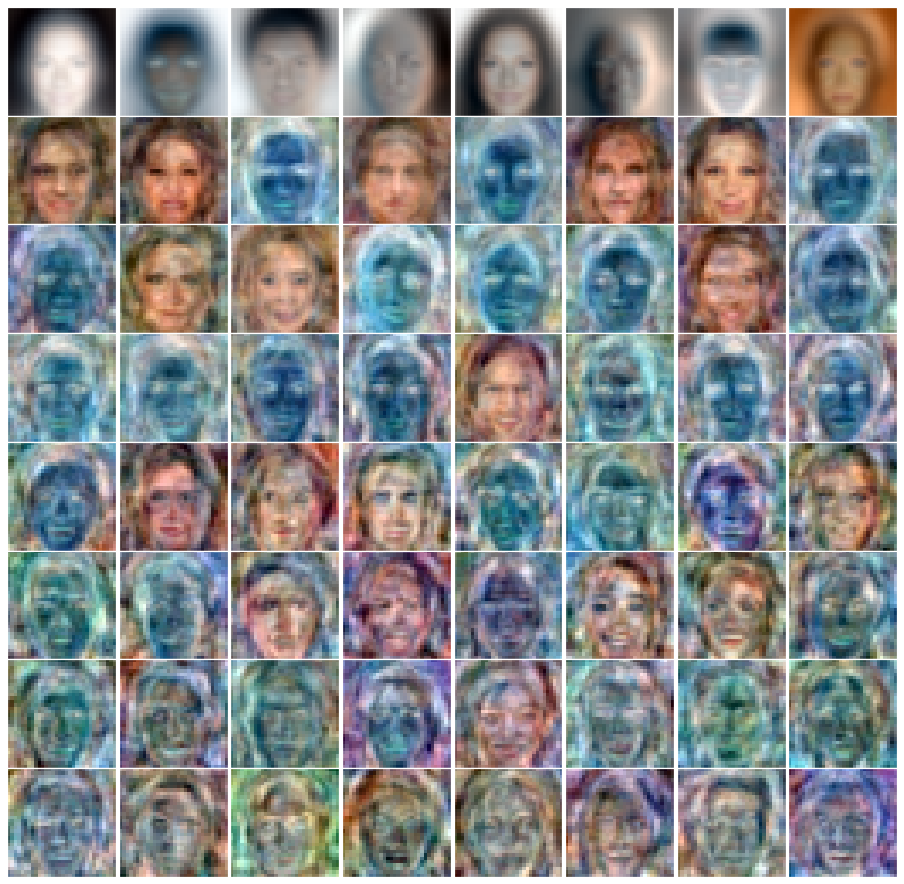}}
    ~
    \subfigure[Thresholded Gaussian+Gamma]{\includegraphics[width=0.46\textwidth]{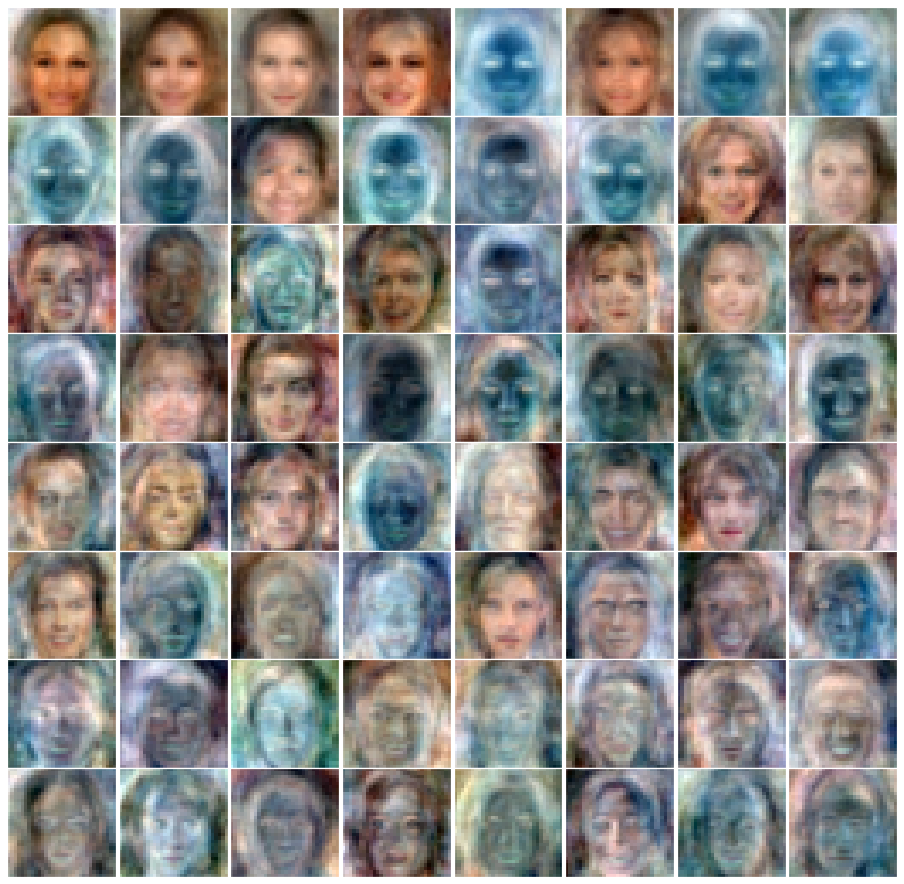}}
    ~
    \caption{\textbf{4096 dimensional latent space}. Sampled entries from an estimated dictionary using an inference network trained with a DNN generator on CelebA. Entries are shown in descending order of Frobenius norm magnitude, with each row taking linearly spaced entries from the full dictionary. The sparse prior is set so to encourage 10\% of latent features to be non-zero.}
\end{figure*}

%%%%%%%%%%%%%%%%%%%%%%%%%%%%%%%%%%%%%%%%%%%%%%%%%%%%%%%%%%%%%%%%%%%%%%%%%%%%%%%
%%%%%%%%%%%%%%%%%%%%%%%%%%%%%%%%%%%%%%%%%%%%%%%%%%%%%%%%%%%%%%%%%%%%%%%%%%%%%%%

\end{document}